%% file: main.tex
\documentclass[10pt,twocolumn,letterpaper]{article}

\usepackage[pagenumbers]{wacv} 


\definecolor{wacvblue}{rgb}{0.21,0.49,0.74}
\usepackage[pagebackref,breaklinks,colorlinks,allcolors=wacvblue]{hyperref}
\usepackage{multirow}
\usepackage[table]{xcolor}
\usepackage{tikz}
\usetikzlibrary{positioning, arrows.meta, shapes.geometric, fit, backgrounds, calc}
\usepackage[accsupp]{axessibility}  
\usetikzlibrary{shadows}
\def\wacvPaperID{2016} 
\def\confName{WACV}
\def\confYear{2027}

\title{Don't Replace, Complement: A Detector-Agnostic Subspace Signal \\for Near-OOD Detection}

\author{
Wonseok Shin \quad Songkuk Kim\thanks{Corresponding author.}\\
Yonsei University, Republic of Korea\\
{\tt\small \{wonseok.shin,songkuk\}@yonsei.ac.kr}
}

\begin{document}
\maketitle
\input{sec/0_abstract}    
\input{sec/1_intro}
\input{sec/2_related_work}
\input{sec/3_method}
\input{sec/4_experimental_setup}
\input{sec/5_result}
\input{sec/6_analysis}
\input{sec/7_conclusion}
\newpage
{
    \small
    \bibliographystyle{ieeenat_fullname}
    \bibliography{main}
}

\appendix
\onecolumn
\begin{center}
  {\Large\bfseries Supplementary Material\par}
  \vspace{0.6em}
  {\large Don't Replace, Complement: A Detector-Agnostic Subspace Signal for Near-OOD Detection\par}
\end{center}
\medskip
\input{sec/99_appendix}

\end{document}

%% file: sec/0_abstract.tex
\begin{abstract}
Post-hoc out-of-distribution (OOD) detection is mature on far-OOD yet remains weak on near-OOD. These two settings differ in their use of class-characteristic feature directions: far-OOD inputs largely fail to activate those of their predicted class, whereas near-OOD inputs activate them much as in-distribution samples do. Because the axis separating far-OOD leaves near-OOD almost indistinguishable from ID, a score optimized for one setting is misaligned with the other; where a detector already saturates far-OOD, added near-OOD sensitivity is bought partly against it. This divergence motivates a targeted near-OOD signal designed to complement existing scores. We propose PLUS (Per-class Low-dimensional sUbspace Signal). For each class, PLUS selects the $m$ most class-active neurons from a sparse autoencoder, orthonormalizes the corresponding decoder directions to form a low-dimensional subspace, and scores an input by its per-class whitened squared Mahalanobis distance $z$ within the subspace of its predicted class. Designed as a dedicated complement rather than a standalone detector, the signal $z$ is fused additively with any calibrated base score at a single, fixed weight. This weight is a single global constant: no value is adjusted per backbone, per OOD dataset, or per deployment, and all calibration statistics are estimated from ID validation data alone. When applied to 18 existing detectors spanning diverse method families, $z$ consistently improves near-OOD AUROC in all 126 detector--backbone combinations. Far-OOD performance is broadly maintained or improved, with only marginal declines where the base score is near-optimal, making $z$ a robust, detector-agnostic complement across backbones.
\end{abstract}

%% file: sec/1_intro.tex
\definecolor{cPlus}{RGB}{224,234,248}
\definecolor{cPlusEdge}{RGB}{46,89,132}
\definecolor{cClf}{RGB}{252,244,215}
\definecolor{cClfEdge}{RGB}{160,130,40}
\definecolor{cFinal}{RGB}{226,242,226}
\definecolor{cFinalEdge}{RGB}{70,130,70}
\definecolor{cBase}{RGB}{250,228,228}
\definecolor{cBaseEdge}{RGB}{164,74,74}
\definecolor{cSae}{RGB}{240,232,219}
\definecolor{cSaeEdge}{RGB}{150,118,76}
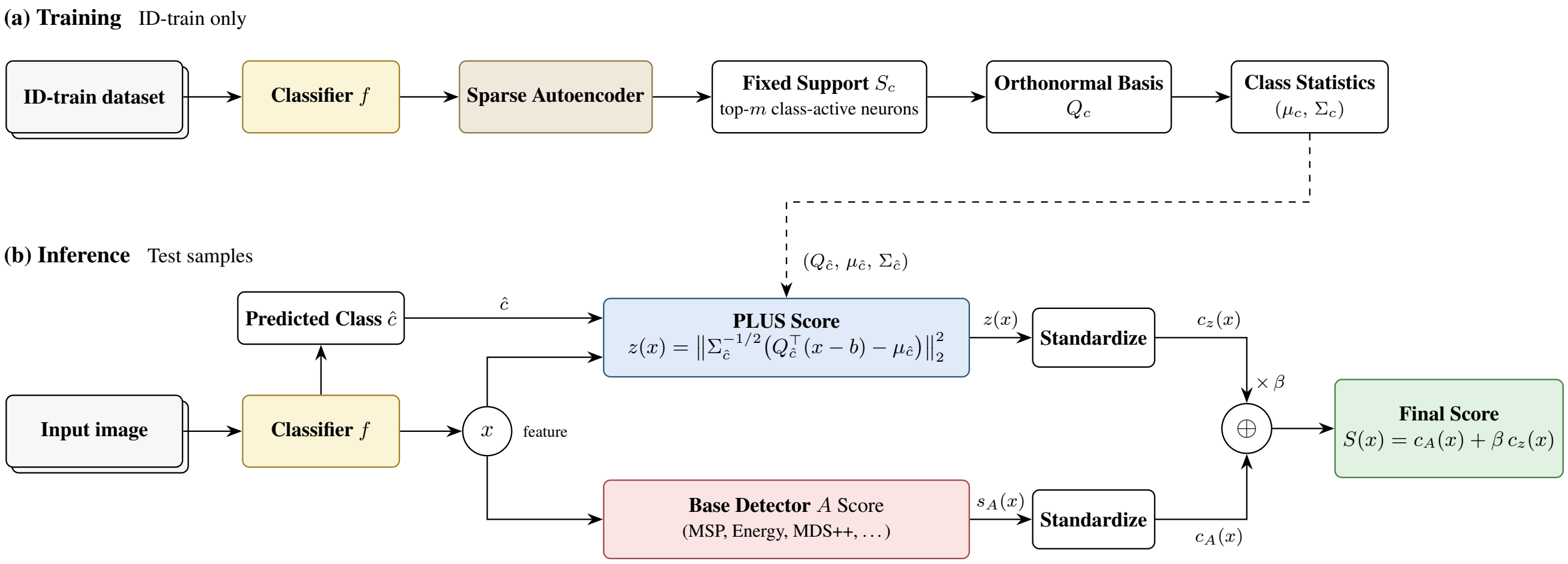
\begin{figure*}[!t]
\centering
\resizebox{0.98\textwidth}{!}{%
\begin{tikzpicture}[
    module/.style={draw=black, fill=white, rectangle, rounded corners=3pt, minimum height=1.1cm, minimum width=2.4cm, align=center, font=\small, semithick},
    pmodule/.style={module, draw=cPlusEdge, fill=cPlus},
    data/.style={draw=black, fill=gray!6, rectangle, rounded corners=3pt, minimum height=1.1cm, minimum width=2.7cm, align=center, font=\small\bfseries, semithick, copy shadow={shadow xshift=2.5pt, shadow yshift=-2.5pt, fill=gray!6, draw=black}},
    op/.style={draw=black, circle, semithick, minimum size=0.75cm, font=\large, fill=white},
    arrow/.style={-{Stealth[length=2.4mm]}, semithick},
    darrow/.style={-{Stealth[length=2.4mm]}, semithick, dashed},
    mathlabel/.style={font=\footnotesize},
    subtitle/.style={font=\normalsize\bfseries, anchor=south west}
]
\node (idtrain) [data] {ID-train dataset};
\node (clft) [module, draw=cClfEdge, fill=cClf, right=0.9cm of idtrain] {\textbf{Classifier} $f$};
\node (sae) [module, draw=cSaeEdge, fill=cSae, right=0.9cm of clft] {\textbf{Sparse Autoencoder}};
\node (supp) [module, right=0.9cm of sae] {\textbf{Fixed Support} $S_c$\\ \footnotesize top-$m$ class-active neurons};
\node (qr) [module, right=0.9cm of supp] {\textbf{Orthonormal Basis} \\$Q_c$};
\node (stats) [module, right=0.9cm of qr] {\textbf{Class Statistics}\\ \footnotesize $(\mu_c,\, \Sigma_c)$};
\draw [arrow] (idtrain) -- (clft);
\draw [arrow] (clft) -- (sae);
\draw [arrow] (sae) -- (supp);
\draw [arrow] (supp) -- (qr);
\draw [arrow] (qr) -- (stats);
\node [subtitle] at ($(idtrain.north west)+(-0.15,0.35)$) {(a) Training \enspace {\normalfont\small ID-train only}};
\node (input) [data, below=4.0cm of idtrain] {Input image};
\node (clfi) [module, draw=cClfEdge, fill=cClf, right=0.9cm of input] {\textbf{Classifier} $f$};
\node (feat) [op, right=0.95cm of clfi, inner sep=2pt, font=\normalsize] {$x$};
\node [right=0.04cm of feat, font=\scriptsize] {feature};
\draw [arrow] (input) -- (clfi);
\draw [arrow] (clfi) -- (feat);
\node (plus) [pmodule, above right=0.55cm and 1.5cm of feat, minimum height=1.2cm, minimum width=5.6cm] {\textbf{PLUS Score}\\ $z(x) = \big\lVert \Sigma_{\hat{c}}^{-1/2} \big( Q_{\hat{c}}^\top (x - b) - \mu_{\hat{c}} \big) \big\rVert_2^2$};
\node (stdZ) [module, right=0.95cm of plus, minimum width=1.7cm, minimum height=0.9cm] {\textbf{Standardize}};
\node (base) [module, draw=cBaseEdge, fill=cBase, minimum height=1.2cm, minimum width=5.6cm] at ($(plus.center |- feat.center)+(0,-1.35)$) {\textbf{Base Detector} $A$ Score\\ \footnotesize (MSP, Energy, MDS++, \dots)};
\node (stdA) [module, minimum width=1.7cm, minimum height=0.9cm] at (stdZ.center |- base.center) {\textbf{Standardize}};
\node (pred) [module, minimum height=0.8cm, minimum width=1.9cm] at ($(clfi.center |- plus.west)+(0,0.3)$) {\textbf{Predicted Class} $\hat{c}$};
\draw [arrow] (clfi.north) -- (pred.south);
\draw [arrow] (pred.east) -- node[midway, above, mathlabel] {$\hat{c}$} ($(plus.west)+(0,0.3)$);
\draw [arrow] (feat) |- ($(plus.west)+(0,-0.3)$);
\draw [arrow] (feat) |- (base.west);
\draw [arrow] (plus) -- node[midway, above, mathlabel] {$z(x)$} (stdZ);
\draw [arrow] (base) -- node[midway, above, mathlabel] {$s_A(x)$} (stdA);
\node (oplus) [op] at ($(stdZ.east)!0.5!(stdA.east) + (1.4cm, 0)$) {$\oplus$};
\draw [arrow] (stdZ.east) -| node[pos=0.35, above, mathlabel] {$c_z(x)$} node[pos=0.85, right, mathlabel] {$\times\, \beta$} (oplus.north);
\draw [arrow] (stdA.east) -| node[pos=0.35, below, mathlabel] {$c_A(x)$} (oplus.south);
\node (final) [module, draw=cFinalEdge, fill=cFinal, minimum height=1.5cm, minimum width=3.4cm, right=0.95cm of oplus] {\textbf{Final Score}\\ $S(x) = c_A(x) + \beta\, c_z(x)$};
\draw [arrow] (oplus) -- (final);
\node [subtitle] at ($(input.west |- plus.north)+(-0.15,0.35)$) {(b) Inference \enspace {\normalfont\small Test samples}};
\draw [darrow] (stats.south) -- ++(0,-1.05) -| (plus.north);
\node [mathlabel, anchor=west] at ($(plus.north)+(0.12,0.55)$) {$(Q_{\hat{c}},\, \mu_{\hat{c}},\, \Sigma_{\hat{c}})$};
\end{tikzpicture}%
}
\vspace{-0.4em}
\caption{\textbf{Overview of PLUS.} \emph{(a) Training:} Using only ID training data, a TopK-SAE is trained on backbone features. For each class $c$, a fixed support $S_c$ of its $m$ most active neurons is identified, orthonormalized into a basis $Q_c$, and characterized by subspace statistics $(\mu_c, \Sigma_c)$. \emph{(b) Inference:} Given an input image, the classifier produces the $\ell_2$-normalized feature $x$ and the predicted class $\hat{c}$. PLUS then computes the whitened deviation $z(x)$ within the predicted class's subspace. Both the base detector score and $z(x)$ are standardized using correctly classified ID validation negatives and additively fused with a single fixed weight $\beta = 0.5$.}
\label{fig:pipeline}
\end{figure*}

\section{Introduction}
\label{sec:intro}
Deploying machine learning models in open-world environments requires the reliable identification of out-of-distribution (OOD) inputs to ensure safety and trustworthiness. Compared to alternative paradigms, post-hoc OOD detection is particularly well-suited for real-world deployment, as it operates directly on pre-trained classifiers without requiring model retraining, architectural modifications, or exposure to auxiliary OOD data. While these post-hoc methods are now highly effective at detecting far-OOD inputs---samples semantically distant from the training data---they exhibit degraded performance on near-OOD inputs, which are visually and semantically proximate to the in-distribution (ID) classes. 

This performance gap is a structural consequence of how features are activated. Far-OOD inputs activate fewer of their class-specific feature directions, making them easily distinguishable via global feature summaries such as density, residual energy, or classifier confidence. In contrast, near-OOD inputs lie directly on the in-distribution manifold and activate class-specific features in a manner closely resembling ID inputs. As a result, the standard global feature axes that successfully isolate far-OOD data leave near-OOD inputs almost indistinguishable from ID data, as they inherently fail to capture the fine-grained internal structure of individual classes.

To overcome this structural limitation, we propose a different approach. Using a sparse autoencoder (SAE), we identify the feature neurons most characteristic of each class and treat their corresponding directions as the axes of a compact, per-class subspace. An input is then evaluated based on its projection into the subspace of its predicted class: a genuine ID sample naturally aligns with the dense regions of the training data, whereas a near-OOD sample exhibits measurable deviation within this subspace. By explicitly capturing this within-class structure, we derive a signal $z$ that is informative precisely where global scores lack discriminative power, providing an effective mechanism for near-OOD detection.

A further challenge is that relying on a single standalone score typically forces an inherent trade-off between near- and far-OOD detection. Rather than attempting to resolve this tension within yet another standalone detector, we introduce \textbf{PLUS} (Per-class Low-dimensional sUbspace Signal). PLUS is designed as a complement that can be seamlessly attached to any existing post-hoc detector, additively integrating our near-OOD signal $z$ via a single, fixed weight. In doing so, PLUS preserves the strong far-OOD capabilities inherent to the base detector while markedly boosting its near-OOD sensitivity---all without network retraining or architectural modifications, and with every hyperparameter held as a single global constant and all calibration statistics estimated from ID validation data alone. Our primary contributions are as follows:

\vspace{0.7em}
\begin{enumerate}
\item
  \textbf{Attributing the Near-OOD Gain to Subspace and Metric.} A subspace$\times$metric ablation on the three main backbones (Section~\ref{sec:ablation_metric}) reveals that the two components of $z$ play distinct roles. The subspace is essential: applying the same per-class metric within a random $16$-dimensional subspace yields no near-OOD gain over MDS++. The per-class metric is required to complete this gain: within our subspace, a shared or isotropic metric recovers only part of the improvement, and a shared metric recovers almost none on ResNet-50. The near-OOD signal thus requires modeling each class's own anisotropy along its most active directions.
\vspace{0.4em}
\item
  \textbf{A Structural Use of a Sparse Autoencoder for Near-OOD.} PLUS uses an SAE structurally rather than as a feature score: each class's most active neurons define a fixed, low-dimensional subspace in which $z$ is measured (Section~\ref{sec:method}). Fixing the support is essential, because raw SAE activations change support from image to image and provide no consistent per-class coordinate system (Section~\ref{sec:method_subspace}). The choice of neurons also matters: $16$ randomly chosen SAE neurons yield essentially no near-OOD gain, and across all seven backbones, the SAE subspace is more consistent than per-class PCA, with a smaller worst-case shortfall (Section~\ref{sec:ablation_metric}, Appendix~\ref{app:sae-vs-pca}).
\vspace{0.4em}
\item
  \textbf{A Detector-Agnostic Complementary Protocol.} Fused additively via a single fixed weight, $z$ improves the near-OOD AUROC of all 18 evaluated detectors across seven diverse architectures while broadly maintaining far-OOD performance (Section~\ref{sec:result_main}, Appendix~\ref{app:fullresults}). This improvement transcends a generic ensemble effect: incorporating $z$ into any of the 714 evaluated two-detector ensembles consistently yields further improvements in near-OOD AUROC across all cases (Section~\ref{sec:ensemble}). Furthermore, all hyperparameters are defined as single global constants shared across backbones and datasets, with calibration frozen on ID validation negatives.
\end{enumerate}

%% file: sec/2_related_work.tex
\section{Related Work}
\label{sec:related_work}

\noindent \textbf{Post-hoc OOD Detectors.} Post-hoc OOD detection typically leverages representations from a frozen classifier without requiring model retraining. Initial approaches primarily focus on the output-space confidence. Methods such as MSP~\cite{hendrycks2017msp}, MaxLogit~\cite{hendrycks2022scaling}, and Energy~\cite{liu2020energy} utilize the maximum softmax probability, raw logits, or free energy to flag inputs when the network is unconfident. Extensions like GEN~\cite{liu2023gen} and KL-Matching~\cite{hendrycks2022scaling} further refine this by measuring entropy or comparing posteriors against class-conditional means.

To overcome the inherent overconfidence of softmax outputs, subsequent research has shifted toward feature-space distance and density estimation. Methods such as Mahalanobis~\cite{lee2018mahalanobis} and RMDS~\cite{ren2021rmds} fit class-conditional Gaussians to score samples based on their deviation from typical ID feature distributions. Recent variants, including MDS++/RMDS++~\cite{mueller2025mahalanobispp} and MahaVar~\cite{kim2026mahavar}, enhance stability by normalizing features or incorporating distance-variance terms. Non-parametric approaches such as KNN~\cite{sun2022knn}, SHE~\cite{zhang2023she}, and NNGuide~\cite{park2023nnguide} similarly identify OOD inputs by measuring their deviation from stored ID prototypes or neighborhood densities.

Another prominent line of work examines the geometric structure of the feature space. Building upon NuSA~\cite{cook2020nusa}, ViM~\cite{wang2022vim} scores the residual norm of features outside the principal ID subspace. Motivated by neural collapse, NECO~\cite{ammar2024neco} instead scores the relative norm within this subspace. Kernel PCA~\cite{fang2024kpca} extends ViM's reconstruction-error view to a non-linear, kernel-induced subspace. Other detectors rely on different cues: fDBD~\cite{liu2024fdbd} measures the distance to decision boundaries, and GradNorm~\cite{huang2021gradnorm} scores the norm of the gradient. Complementary to these scoring functions, shaping techniques such as ReAct~\cite{sun2021react}, ASH~\cite{djurisic2023ash}, and DICE~\cite{sun2022dice} truncate, rescale, or sparsify activations or weights to amplify the margin between ID and OOD samples.

While highly effective, the operative signal across all these families is inherently global: they measure general atypicality, distance to the overall ID feature cloud, or deviation from a class-agnostic subspace shared by all classes (e.g., ViM~\cite{wang2022vim} or Kernel PCA~\cite{fang2024kpca}). Consequently, they are structurally blind to near-OOD inputs, which lie on the ID manifold and mimic the typical features of their predicted class. Our method is designed specifically to address this gap. Rather than proposing another standalone detector, PLUS introduces a targeted, per-class deviation signal designed to be additively fused with any of the existing global scores mentioned above.

\vspace{0.7em}

\noindent \textbf{Sparse Autoencoders.}
Sparse Autoencoders (SAEs) recently gained prominence in language-model interpretability as a dictionary learning tool. They employ an overcomplete, sparsely activated architecture to decompose dense representations into monosemantic (interpretable) feature directions~\cite{bricken2023monosemanticity,cunningham2023sparse}. Beyond the standard ReLU architecture, modern variants like TopK~\cite{gao2024topksae}, Gated~\cite{rajamanoharan2024gated}, and JumpReLU~\cite{rajamanoharan2024jumprelu} SAEs balance reconstruction fidelity and sparsity differently, enabling dictionaries to scale up to millions of neurons~\cite{templeton2024scaling}.

However, whether SAEs genuinely disentangle human-interpretable concepts remains widely debated. Empirical evaluations indicate that SAE features only partially align with true causal directions~\cite{makelov2024principled}, and models trained with different random seeds often yield substantially different dictionaries~\cite{paulo2025different}. This lack of stable semantic alignment is even more pronounced in computer vision, where the evidence for human-level interpretability is still preliminary.

In the context of OOD detection, Oh et al.~\cite{oh2026sparsity} recently utilized SAEs by scoring samples based directly on the activation typicality of their latent neurons. PLUS, however, adopts a fundamentally different perspective. We make no claims of human-level semantic interpretability; instead, PLUS treats the SAE purely as a geometric tool. We identify the neurons most characteristically active for each class and use their corresponding decoder weights to construct a fixed, low-dimensional projection basis. By exploiting the overcomplete dictionary as a source of class-specific coordinate directions, PLUS bypasses the often-debated pursuit of semantic interpretability, leveraging the SAE for robust structural isolation.

%% file: sec/3_method.tex
\section{Method}
\label{sec:method}
Our method, PLUS, constructs a low-dimensional per-class projection basis from the decoder of a sparse autoencoder (SAE). At inference time, it produces a single scalar $z(x)$ representing the whitened deviation of an input within the subspace of its predicted class. This signal is then fused additively with any calibrated base detector score at a single fixed weight. Figure~\ref{fig:pipeline} provides an overview of PLUS.

\subsection{Preliminaries}
Let $f$ denote a frozen, pretrained classifier and let $x' \in \mathbb{R}^d$ be its penultimate-layer feature. To remove magnitude bias, we operate on the $\ell_2$-normalized feature $x = x' / \Vert x' \Vert_2$. The classifier's prediction $\hat{c}(x) \in \{1, \ldots, C\}$ is used directly.

To construct our per-class subspaces, we extract an overcomplete dictionary of $p \gg d$ unit-norm directions (referred to as \emph{neurons}) from the normalized ID-train features $x$ using a TopK sparse autoencoder (SAE)~\cite{gao2024topksae}. The SAE encodes the input into a sparse activation vector with at most $k$ non-zero entries:
\begin{equation}
h(x) = \mathrm{ReLU}\big(\mathrm{TopK}_k\big(W_{\text{enc}}(x - b) + b_{\text{enc}}\big)\big) \in \mathbb{R}_{\geq 0}^{p}.
\end{equation}
It then reconstructs the feature via a linear decoder $\hat{x} = W_{\text{dec}}\, h(x) + b$, where $W_{\text{dec}} = [w_1, \ldots, w_p] \in \mathbb{R}^{d \times p}$ forms the dictionary. Trained unsupervised via standard reconstruction and auxiliary losses, the SAE provides the ID-train activations $h(x)$ and the decoder dictionary $\{w_j\}_{j=1}^p$ needed to establish our per-class supports.

\subsection{Subspace Construction}
\label{sec:method_subspace}

Raw TopK-SAE latents activate a different support on every image; within a single class, the top-$k$ active-neuron sets of two images share only a small fraction of neurons (intra-class support Jaccard $\approx 0.09$ on ViT-B/16, versus 1 by construction for the fixed support). A per-image support implies a shifting coordinate system, rendering any class-conditional distribution undefined. To make the geometric object ``distribution of class $c$ in its subspace'' well-posed, we must establish a consistent projection frame. For each class $c$, we fix the support $S_c \subset \{1, \ldots, p\}$ ($|S_c| = m = 16$) as the indices of the $m$ neurons maximizing $\bar{h}_j(c) = \frac{1}{n_c} \sum_{i : y_i = c} h_j(x_i)$, where $n_c$ is the number of ID-train samples for class $c$. Computed over post-TopK activations, this selection favors the neurons that fire most frequently and strongly for class $c$.

To build a well-defined orthonormal basis for this support, we collect the corresponding decoder columns $W_{\text{dec}}[:, S_c] \in \mathbb{R}^{d \times m}$ and apply a thin QR factorization with a fixed sign convention:
\begin{equation}
\begin{aligned} 
W_{\text{dec}}[:, S_c] &= Q_c R_c, \quad Q_c \in \mathbb{R}^{d \times m}, \\ 
Q_c^\top Q_c &= I_m, \quad \text{diag}(R_c) > 0. 
\end{aligned}
\end{equation}
$Q_c$ spans the same subspace as the selected neurons but in an orthonormal frame, ensuring that correlations among dictionary directions do not distort the distance metric. With the sign convention $\text{diag}(R_c) > 0$, the factorization is unique, making $Q_c$ a purely deterministic function of $S_c$ and the shared dictionary.

\subsection{The Signal \texorpdfstring{$z$}{z}}

Given the fixed orthonormal basis $Q_c \in \mathbb{R}^{d \times m}$ for each class $c$, the empirical mean and covariance of the centered, projected ID-train features in the $m$-dimensional subspace are given by:
\begin{equation}
\begin{gathered}
\mu_c = \frac{1}{n_c} \sum_{i : y_i = c} Q_c^\top (x_i - b) \in \mathbb{R}^m, \\
\Sigma_c = \frac{1}{n_c} \sum_{i : y_i = c} u_{i,c} u_{i,c}^\top + \epsilon I_m,
\end{gathered}
\end{equation}
where $u_{i,c} = Q_c^\top (x_i - b) - \mu_c$, $b \in \mathbb{R}^d$ is the SAE decoder bias, and $\epsilon = 10^{-3}$ is a ridge regularization parameter fixed across all classes and backbones. The low dimensionality ($m = 16$) ensures that $\Sigma_c$ is well-conditioned, avoiding the singularity inherent to full $d$-dimensional per-class covariances at ImageNet scales.

At test time, given a normalized feature $x$ and predicted class $\hat{c} = \hat{c}(x)$, the centered feature is projected onto $Q_{\hat{c}}$ to evaluate the whitened squared Mahalanobis distance within the corresponding subspace:
\begin{equation} 
z(x) = \big\lVert \Sigma_{\hat{c}}^{-1/2} \big( Q_{\hat{c}}^\top (x - b) - \mu_{\hat{c}} \big) \big\rVert_2^2.
\end{equation}
Here, $Q_{\hat{c}}$ spans the directions in which class $\hat{c}$ concentrates its activations, while $\Sigma_{\hat{c}}^{-1/2}$ weights each deviation by that class's own variability, amplifying displacements along directions where the class varies little, and thereby quantifies atypicality along the coordinates most relevant to the predicted class. $z$ discards the component of $x$ orthogonal to the subspace, measuring instead how atypical the in-subspace coordinates are relative to the class's own ID distribution.

\subsection{Calibration and Fusion}

Let $s_A(x)$ denote the score of any base detector $A$, with its sign oriented so that larger indicates more OOD (for detectors with the opposite native convention, e.g., MSP, we use $-s_A$; the orientation is a fixed property of each method). Because $s_A$ and $z$ live on unrelated scales, both are standardized on the correctly-classified ID-validation negatives using a single global mean and standard deviation per channel, frozen once:
\begin{equation}
c_A(x) = \frac{s_A(x) - \nu_A}{\tau_A}, \quad
c_z(x) = \frac{z(x) - \nu_z}{\tau_z},
\end{equation}
where $(\nu_\bullet, \tau_\bullet)$ are the mean and standard deviation of the corresponding score over these negatives, computed once and frozen.

Standardization is a monotone affine map, leaving each channel's individual AUROC unchanged; its role is to express both channels in common ID-standard-deviation units, so that the fixed weight below is interpretable as a channel ratio. The deployed score combines these complementary mechanisms via:
\begin{equation}
S(x) = c_A(x) + \beta c_z(x), \quad \beta = 0.5.
\end{equation}
We fix $\beta = 0.5$ as an \textit{a priori} base-favored choice (a 2:1 ratio reflecting higher trust in the established base detector), which is neither fitted, tuned, nor selected by any algorithmic rule on any split. We emphasize that $\beta = 0.5$ is a deployment default, not a claim of optimality. The sensitivity analysis (Section \ref{sec:beta}) shows the trade-off is backbone-dependent: near-OOD gains saturate around $\beta \approx 0.5$ on well-conditioned backbones but continue rising beyond it on others, while smaller weights (e.g., $\beta = 0.25$) retain most of the near-OOD gain at a lower far-OOD cost. A practitioner with deployment-specific priorities, or validation data of their own, may therefore prefer a different operating point on this curve; all results reported in this paper use the single fixed $\beta = 0.5$.

%% file: sec/4_experimental_setup.tex
\begin{table*}[t]
\centering
\setlength{\tabcolsep}{2.5pt}
\renewcommand{\arraystretch}{1.2}
\resizebox{\textwidth}{!}{%
\begin{tabular}{@{}l *{18}{c}@{}}
\toprule
\multirow{3}{*}{\textbf{Method}} & \multicolumn{6}{c}{\textbf{ViT-B/16}} & \multicolumn{6}{c}{\textbf{ResNet-50}} & \multicolumn{6}{c}{\textbf{Swin-T}} \\
\cmidrule(lr){2-7}\cmidrule(lr){8-13}\cmidrule(lr){14-19}
 & \multicolumn{2}{c}{\textbf{Near-OOD}} & \multicolumn{2}{c}{\textbf{Far-OOD}} & \multicolumn{2}{c}{\textbf{Overall}} & \multicolumn{2}{c}{\textbf{Near-OOD}} & \multicolumn{2}{c}{\textbf{Far-OOD}} & \multicolumn{2}{c}{\textbf{Overall}} & \multicolumn{2}{c}{\textbf{Near-OOD}} & \multicolumn{2}{c}{\textbf{Far-OOD}} & \multicolumn{2}{c}{\textbf{Overall}} \\
\cmidrule(lr){2-3}\cmidrule(lr){4-5}\cmidrule(lr){6-7}\cmidrule(lr){8-9}\cmidrule(lr){10-11}\cmidrule(lr){12-13}\cmidrule(lr){14-15}\cmidrule(lr){16-17}\cmidrule(lr){18-19}
 & \textbf{AUROC~$\uparrow$} & \textbf{FPR95~$\downarrow$} & \textbf{AUROC~$\uparrow$} & \textbf{FPR95~$\downarrow$} & \textbf{AUROC~$\uparrow$} & \textbf{FPR95~$\downarrow$} & \textbf{AUROC~$\uparrow$} & \textbf{FPR95~$\downarrow$} & \textbf{AUROC~$\uparrow$} & \textbf{FPR95~$\downarrow$} & \textbf{AUROC~$\uparrow$} & \textbf{FPR95~$\downarrow$} & \textbf{AUROC~$\uparrow$} & \textbf{FPR95~$\downarrow$} & \textbf{AUROC~$\uparrow$} & \textbf{FPR95~$\downarrow$} & \textbf{AUROC~$\uparrow$} & \textbf{FPR95~$\downarrow$} \\
\midrule
MSP & 73.52 & 81.85 & 86.04 & 51.69 & 81.03 & 63.75 & 76.02 & 65.68 & 85.23 & 51.45 & 81.55 & 57.14 & 76.75 & 71.06 & 86.30 & 49.16 & 82.48 & 57.92 \\
\rowcolor{gray!9}\textbf{MSP + PLUS} & \textbf{78.13} & \textbf{68.82} & \textbf{88.44} & \textbf{43.86} & \textbf{84.32} & \textbf{53.84} & \textbf{79.30} & \textbf{63.38} & \textbf{89.64} & \textbf{40.88} & \textbf{85.50} & \textbf{49.88} & \textbf{80.02} & \textbf{64.33} & \textbf{88.55} & \textbf{42.96} & \textbf{85.14} & \textbf{51.51} \\
MaxLogit & 68.30 & 92.25 & 83.54 & 79.23 & 77.45 & 84.44 & 76.46 & 67.85 & 89.58 & 38.20 & 84.33 & 50.06 & 75.66 & 80.76 & 84.80 & 67.81 & 81.14 & 72.99 \\
\rowcolor{gray!9}\textbf{MaxLogit + PLUS} & \textbf{75.65} & \textbf{84.39} & \textbf{87.44} & \textbf{56.73} & \textbf{82.73} & \textbf{67.80} & \textbf{80.86} & \textbf{63.19} & \textbf{92.98} & \textbf{29.22} & \textbf{88.13} & \textbf{42.81} & \textbf{79.79} & \textbf{72.02} & \textbf{87.79} & \textbf{51.50} & \textbf{84.59} & \textbf{59.70} \\
Energy & 62.41 & 93.19 & 78.98 & 85.35 & 72.35 & 88.48 & 75.89 & 68.56 & 89.47 & 38.39 & 84.04 & 50.46 & 73.23 & 83.31 & 81.32 & 75.59 & 78.08 & 78.68 \\
\rowcolor{gray!9}\textbf{Energy + PLUS} & \textbf{73.27} & \textbf{89.51} & \textbf{85.86} & \textbf{69.29} & \textbf{80.82} & \textbf{77.38} & \textbf{80.71} & \textbf{63.57} & \textbf{92.95} & \textbf{29.01} & \textbf{88.05} & \textbf{42.83} & \textbf{78.95} & \textbf{75.61} & \textbf{86.26} & \textbf{58.87} & \textbf{83.34} & \textbf{65.57} \\
GEN & 76.30 & 70.78 & 91.35 & 32.23 & 85.33 & 47.65 & 76.85 & 65.29 & 89.77 & 35.62 & 84.60 & 47.49 & 78.97 & 63.62 & 90.98 & 32.59 & 86.17 & 45.00 \\
\rowcolor{gray!9}\textbf{GEN + PLUS} & \textbf{79.68} & \textbf{65.87} & \textbf{92.13} & \textbf{30.53} & \textbf{87.15} & \textbf{44.66} & \textbf{81.11} & \textbf{62.18} & \textbf{92.91} & \textbf{28.17} & \textbf{88.19} & \textbf{41.78} & \textbf{81.59} & \textbf{60.81} & \textbf{91.79} & \textbf{30.42} & \textbf{87.71} & \textbf{42.57} \\
KL-Matching & 74.41 & 77.25 & 87.70 & 48.45 & 82.39 & 59.97 & 76.63 & 72.53 & 87.59 & 46.81 & 83.21 & 57.10 & 76.11 & 75.46 & 88.32 & 43.33 & 83.44 & 56.18 \\
\rowcolor{gray!9}\textbf{KL-Matching + PLUS} & \textbf{77.04} & \textbf{70.52} & \textbf{88.92} & \textbf{40.57} & \textbf{84.17} & \textbf{52.55} & \textbf{78.51} & \textbf{65.81} & \textbf{89.40} & \textbf{37.95} & \textbf{85.04} & \textbf{49.09} & \textbf{78.28} & \textbf{67.89} & \textbf{89.43} & \textbf{38.52} & \textbf{84.97} & \textbf{50.27} \\
ReAct & 69.26 & 84.49 & 85.69 & 53.93 & 79.12 & 66.15 & 77.38 & 66.74 & 93.67 & 26.31 & 87.15 & 42.48 & 75.64 & 72.66 & 88.23 & 43.01 & 83.19 & 54.87 \\
\rowcolor{gray!9}\textbf{ReAct + PLUS} & \textbf{76.15} & \textbf{74.70} & \textbf{89.24} & \textbf{42.91} & \textbf{84.00} & \textbf{55.63} & \textbf{80.87} & \textbf{63.27} & \textbf{94.09} & \textbf{24.35} & \textbf{88.80} & \textbf{39.92} & \textbf{79.82} & \textbf{66.07} & \textbf{90.21} & \textbf{35.96} & \textbf{86.05} & \textbf{48.01} \\
ASH & 53.21 & 94.43 & 51.56 & 96.77 & 52.22 & 95.84 & 78.17 & 63.33 & 95.74 & 19.52 & 88.71 & 37.05 & 46.47 & 95.04 & 44.64 & 94.71 & 45.37 & 94.84 \\
\rowcolor{gray!9}\textbf{ASH + PLUS} & \textbf{70.51} & \textbf{88.88} & \textbf{72.17} & \textbf{89.91} & \textbf{71.51} & \textbf{89.49} & \textbf{81.34} & \textbf{61.32} & \underline{95.61} & \textbf{18.64} & \textbf{89.90} & \textbf{35.71} & \textbf{70.79} & \textbf{87.38} & \textbf{74.64} & \textbf{81.65} & \textbf{73.10} & \textbf{83.94} \\
DICE & 65.42 & 85.32 & 82.57 & 51.31 & 75.71 & 64.91 & 73.08 & 72.44 & 90.95 & 41.81 & 83.80 & 54.06 & 48.46 & 96.87 & 60.68 & 94.13 & 55.79 & 95.23 \\
\rowcolor{gray!9}\textbf{DICE + PLUS} & \textbf{74.63} & \textbf{72.95} & \textbf{88.27} & \textbf{38.21} & \textbf{82.81} & \textbf{52.10} & \textbf{79.37} & \textbf{67.62} & \textbf{93.33} & \textbf{30.61} & \textbf{87.75} & \textbf{45.42} & \textbf{64.43} & \textbf{89.72} & \textbf{76.98} & \textbf{82.35} & \textbf{71.96} & \textbf{85.30} \\
ViM & 77.03 & 73.63 & 92.93 & 29.15 & 86.57 & 46.94 & 72.04 & 71.30 & 92.61 & 24.86 & 84.38 & 43.44 & 75.31 & 74.67 & 93.26 & 24.52 & 86.08 & 44.58 \\
\rowcolor{gray!9}\textbf{ViM + PLUS} & \textbf{79.93} & \textbf{67.98} & \textbf{93.15} & \textbf{26.30} & \textbf{87.87} & \textbf{42.97} & \textbf{77.37} & \textbf{67.56} & \textbf{94.80} & \textbf{21.34} & \textbf{87.83} & \textbf{39.82} & \textbf{78.92} & \textbf{67.34} & \textbf{93.95} & \textbf{24.04} & \textbf{87.94} & \textbf{41.36} \\
KNN & 73.93 & 70.67 & 90.71 & 32.28 & 84.00 & 47.64 & 68.56 & 74.24 & 90.30 & 34.25 & 81.61 & 50.25 & 71.40 & 72.01 & 89.24 & 34.67 & 82.10 & 49.61 \\
\rowcolor{gray!9}\textbf{KNN + PLUS} & \textbf{78.58} & \textbf{67.14} & \textbf{91.72} & \textbf{28.36} & \textbf{86.46} & \textbf{43.87} & \textbf{75.31} & \textbf{70.51} & \textbf{94.14} & \textbf{24.54} & \textbf{86.61} & \textbf{42.93} & \textbf{77.53} & \textbf{67.83} & \textbf{91.62} & \textbf{29.37} & \textbf{85.98} & \textbf{44.76} \\
MDS & 79.03 & 66.11 & 92.65 & 29.94 & 87.20 & 44.41 & 55.44 & 85.40 & 74.10 & 62.97 & 66.64 & 71.94 & 75.20 & 68.81 & 91.51 & 29.97 & 84.98 & 45.51 \\
\rowcolor{gray!9}\textbf{MDS + PLUS} & \textbf{80.94} & \textbf{63.82} & \underline{92.54} & \textbf{28.09} & \textbf{87.90} & \textbf{42.38} & \textbf{68.50} & \textbf{77.73} & \textbf{87.46} & \textbf{44.55} & \textbf{79.88} & \textbf{57.82} & \textbf{78.70} & \textbf{65.27} & \textbf{92.55} & \textbf{28.17} & \textbf{87.01} & \textbf{43.01} \\
RMDS & 79.99 & 65.73 & 92.61 & 28.79 & 87.56 & 43.57 & 76.71 & 65.33 & 86.17 & 41.77 & 82.39 & 51.19 & 78.22 & 66.70 & 92.27 & 28.93 & 86.65 & 44.04 \\
\rowcolor{gray!9}\textbf{RMDS + PLUS} & \textbf{81.47} & \textbf{63.17} & \underline{92.36} & \textbf{27.86} & \textbf{88.00} & \textbf{41.98} & \textbf{80.50} & \textbf{62.19} & \textbf{91.67} & \textbf{32.30} & \textbf{87.20} & \textbf{44.26} & \textbf{80.69} & \textbf{63.77} & \textbf{92.74} & \textbf{27.73} & \textbf{87.92} & \textbf{42.14} \\
MDS++ & 79.87 & 65.63 & 93.29 & 24.93 & 87.92 & 41.21 & 74.71 & 69.05 & 95.13 & 20.37 & 86.96 & 39.85 & 78.92 & 67.66 & 93.67 & 25.17 & 87.77 & 42.16 \\
\rowcolor{gray!9}\textbf{MDS++ + PLUS} & \textbf{81.09} & \textbf{63.87} & \underline{92.26} & \underline{26.58} & \underline{87.79} & \underline{41.50} & \textbf{78.23} & \textbf{66.51} & \textbf{95.45} & \textbf{19.56} & \textbf{88.56} & \textbf{38.34} & \textbf{80.86} & \textbf{65.00} & \underline{93.00} & \underline{25.24} & \textbf{88.14} & \textbf{41.15} \\
RMDS++ & 80.07 & 65.84 & 92.64 & 27.02 & 87.61 & 42.55 & 75.75 & 69.41 & 85.22 & 51.98 & 81.43 & 58.95 & 79.70 & 66.61 & 92.44 & 27.68 & 87.34 & 43.25 \\
\rowcolor{gray!9}\textbf{RMDS++ + PLUS} & \textbf{81.21} & \textbf{63.54} & \underline{91.69} & \underline{28.30} & \underline{87.50} & \textbf{42.40} & \textbf{79.90} & \textbf{65.22} & \textbf{90.88} & \textbf{35.71} & \textbf{86.49} & \textbf{47.52} & \textbf{81.37} & \textbf{63.97} & \underline{91.81} & \underline{27.90} & \textbf{87.63} & \textbf{42.33} \\
GradNorm & 39.28 & 94.71 & 41.75 & 92.65 & 40.76 & 93.47 & 72.96 & 78.88 & 90.25 & 47.91 & 83.34 & 60.30 & 47.58 & 93.82 & 35.47 & 96.61 & 40.32 & 95.50 \\
\rowcolor{gray!9}\textbf{GradNorm + PLUS} & \textbf{59.37} & \textbf{92.45} & \textbf{65.04} & \textbf{90.95} & \textbf{62.77} & \textbf{91.55} & \textbf{78.90} & \textbf{71.36} & \textbf{92.94} & \textbf{35.59} & \textbf{87.33} & \textbf{49.90} & \textbf{64.39} & \textbf{91.83} & \textbf{58.65} & \textbf{94.52} & \textbf{60.95} & \textbf{93.44} \\
SHE & 76.02 & 71.08 & 92.38 & 27.53 & 85.83 & 44.95 & 73.70 & 73.13 & 90.85 & 41.90 & 83.99 & 54.39 & 76.48 & 76.57 & 89.62 & 47.11 & 84.36 & 58.89 \\
\rowcolor{gray!9}\textbf{SHE + PLUS} & \textbf{79.43} & \textbf{66.47} & \textbf{92.72} & \textbf{26.17} & \textbf{87.41} & \textbf{42.29} & \textbf{79.64} & \textbf{66.73} & \textbf{93.67} & \textbf{30.16} & \textbf{88.06} & \textbf{44.79} & \textbf{80.23} & \textbf{69.76} & \textbf{91.20} & \textbf{36.18} & \textbf{86.81} & \textbf{49.61} \\
NNGuide & 57.29 & 91.91 & 74.29 & 74.46 & 67.49 & 81.44 & 77.53 & 65.92 & 94.59 & 25.20 & 87.76 & 41.49 & 60.79 & 91.74 & 81.23 & 65.37 & 73.05 & 75.92 \\
\rowcolor{gray!9}\textbf{NNGuide + PLUS} & \textbf{70.16} & \textbf{85.64} & \textbf{84.62} & \textbf{57.68} & \textbf{78.84} & \textbf{68.87} & \textbf{81.10} & \textbf{62.60} & \textbf{94.83} & \textbf{22.02} & \textbf{89.34} & \textbf{38.25} & \textbf{71.58} & \textbf{84.70} & \textbf{88.04} & \textbf{49.94} & \textbf{81.45} & \textbf{63.84} \\
fDBD & 73.27 & 73.67 & 89.36 & 38.99 & 82.93 & 52.87 & 76.63 & 64.69 & 92.78 & 26.54 & 86.32 & 41.80 & 73.02 & 72.81 & 89.58 & 34.88 & 82.96 & 50.05 \\
\rowcolor{gray!9}\textbf{fDBD + PLUS} & \textbf{78.01} & \textbf{69.29} & \textbf{90.84} & \textbf{33.80} & \textbf{85.71} & \textbf{48.00} & \textbf{80.01} & \textbf{62.70} & \textbf{94.41} & \textbf{23.32} & \textbf{88.65} & \textbf{39.07} & \textbf{77.89} & \textbf{67.66} & \textbf{91.21} & \textbf{31.41} & \textbf{85.88} & \textbf{45.91} \\
\bottomrule
\end{tabular}%
}
\caption{\textbf{PLUS as a detector-agnostic near-OOD complement.} For each of 18 post-hoc baselines on the three standard OpenOOD backbones, the standalone score (plain row) is followed by the $\beta{=}0.5$ fusion (shaded ``base + PLUS'' row). Improvements are shown in \textbf{bold}; an \underline{underline} marks the few metrics where the fusion slightly regresses the standalone (AUROC higher / FPR95 lower is better).}
\label{tab:headline}
\end{table*}

\section{Experimental Setup}
\label{sec:experimental_setup}

\subsection{Benchmark, Backbones, and Baselines}

\noindent\textbf{Benchmark.} 
We evaluate on the OpenOOD v1.5 framework~\cite{zhang2023openood} using ImageNet-1K~\cite{deng2009imagenet} as the in-distribution (ID) dataset. Near-OOD evaluation comprises SSB-hard~\cite{vaze2022ssb} and NINCO~\cite{bitterwolf2023ninco}, while far-OOD evaluation covers iNaturalist~\cite{vanhorn2018inaturalist}, Textures~\cite{cimpoi2014describing}, and OpenImage-O~\cite{wang2022vim}. Performance is evaluated using AUROC and FPR95.

\vspace{0.7em}

\noindent\textbf{Backbones.}
We evaluate the three standard OpenOOD backbones: ViT-B/16~\cite{dosovitskiy2021vit}, ResNet-50~\cite{he2016resnet}, and Swin-T~\cite{liu2021swin}. In addition, to assess generalization across diverse architectural families, we evaluate four further ImageNet backbones (seven in total): ResNet-18~\cite{he2016resnet}, Wide-ResNet50-2~\cite{zagoruyko2016wide}, ConvNeXt-B~\cite{liu2022convnext}, and MaxViT-T~\cite{tu2022maxvit}. Detailed checkpoint sources for all backbones are provided in Appendix~\ref{app:backbone}.

\vspace{0.7em}

\noindent\textbf{Baselines.}
We compare against a comprehensive suite of post-hoc detectors from OpenOOD v1.5 as well as recent state-of-the-art methods: MSP~\cite{hendrycks2017msp}, MaxLogit~\cite{hendrycks2022scaling}, Energy~\cite{liu2020energy}, GEN~\cite{liu2023gen}, KL-Matching~\cite{hendrycks2022scaling}, ReAct~\cite{sun2021react}, ASH~\cite{djurisic2023ash}, DICE~\cite{sun2022dice}, ViM~\cite{wang2022vim}, KNN~\cite{sun2022knn}, MDS~\cite{lee2018mahalanobis}, RMDS~\cite{ren2021rmds}, MDS++/RMDS++~\cite{mueller2025mahalanobispp}, GradNorm~\cite{huang2021gradnorm}, SHE~\cite{zhang2023she}, NNGuide~\cite{park2023nnguide}, and fDBD~\cite{liu2024fdbd}. Our implementations are based on the official OpenOOD v1.5 codebase, while detectors not natively supported by the framework follow the official repositories released by their respective authors.

\subsection{Sparse Autoencoder}

\noindent\textbf{Training and Hyperparameters.}
We employ a TopK sparse autoencoder trained on $\ell_2$-normalized ID-train penultimate features. For each input, the encoder retains only the $k = 32$ largest pre-activations while zeroing the rest, with an AuxK loss reviving dead units to keep the dead-neuron rate negligible. We uniformly fix the number of neurons to $p = 6144$ across all backbones, while specific training procedures follow the original method of \cite{gao2024topksae}.

\vspace{0.7em}

\noindent\textbf{Neuron Demand Across Backbones.} The near-OOD performance response to the neuron count $p$ is largely flat. Sweeping $p$ from $2\times$ to $12\times$ the feature dimension alters the $z$-standalone near-OOD AUROC by at most $0.5$ points on four of the seven backbones, with the response itself exhibiting noise with respect to $p$. The two backbones that do demonstrate a clear preference diverge in opposite directions: ResNet-18 steadily improves as neurons are added, whereas MaxViT-T peaks at the minimal overcomplete capacity, even though both share the exact same feature dimension ($d = 512$) and expansion ratio. We therefore adopt a single fixed $p = 6144$ across all models rather than tuning it per backbone. As detailed in Appendix~\ref{app:psweep}, this single value costs at most $\approx\!1$ point of $z$-standalone near-OOD AUROC relative to the best per-backbone $p$ up to $12\times$, and this gap shrinks further once $z$ is fused with a base detector.

\vspace{0.7em}

\noindent\textbf{Computational cost.} PLUS leaves the base classifier frozen and requires only a one-time SAE fitting on the ID-train features. Training the SAE on cached Swin-T features takes 7--9 minutes on a single NVIDIA RTX A5000, peaking at 3.7\,GiB of memory. At inference, computing $z$ requires one $d{\times}16$ projection and one $16{\times}16$ whitening operation per image. This takes 0.4--0.7\,$\mu$s, representing at most 0.2\% of a typical backbone forward pass. Storing the per-class statistics $\{Q_c, \mu_c, \Sigma_c^{-1/2}\}$ for 1,000 classes requires 32--126\,MiB in fp32, depending on $d$.

\subsection{PLUS Configuration}

For each class, PLUS selects the $m = 16$ most class-active neurons from the SAE, orthonormalizes their decoder directions into a basis $Q_c$ using sign-fixed QR decomposition, and computes the per-class whitened squared Mahalanobis distance $z$ (with a whitening ridge $\epsilon = 10^{-3}$). The supports and the subspace statistics $(\mu_c, \Sigma_c)$ are estimated on a random subset of ID-train images (at most 300 per class). The score $z$ is then standardized and fused with any calibrated base score using a fixed weight $\beta = 0.5$.

%% file: sec/5_result.tex
\section{Results}
\label{sec:result}

\subsection{Results on the Three Main Backbones}
\label{sec:result_main}
Table~\ref{tab:headline} evaluates PLUS across all 18 post-hoc baselines on the three standard OpenOOD backbones, comparing the standalone base detectors against their fusion with $z$ at the fixed weight $\beta = 0.5$. Integrating $z$ improves the near-OOD AUROC in every one of the 54 detector-backbone combinations (18 detectors $\times$ 3 backbones), often by large margins (e.g., $+20.1$ on GradNorm and $+10.9$ on Energy with ViT-B/16, and $+24.3$ on ASH with Swin-T), with FPR95 decreasing correspondingly. Far-OOD and overall AUROC improve in the large majority of cases. The only regressions (underlined) are confined to a few far-OOD metrics on the already far-saturated Mahalanobis-family bases (MDS, RMDS, MDS++, and RMDS++ on ViT-B/16, and MDS++ and RMDS++ on Swin-T) together with ASH on ResNet-50; each AUROC drop is at most $\approx\!1.0$ point.

\subsection{Per-Dataset Results Across Seven Backbones}
\label{sec:cellcounts}

Table~\ref{tab:cellcounts} breaks down the improvement \emph{per dataset} and extends it to all seven backbones, counting how many of the 18 detectors improve versus drop on each OOD dataset once PLUS is added. Near-OOD gains are highly uniform: on the hardest split, SSB-hard, PLUS improves all 18 detectors on every backbone. On NINCO, it improves all 18 on six of the seven backbones (the sole exception being a single detector on Wide-ResNet50-2 with a marginal $-0.23$-point dip).

Far-OOD improves broadly but exposes a per-dataset near/far tension. While iNaturalist and OpenImage-O improve for almost every detector, Textures concentrates the trade-off ($9/9$ improve/drop on ViT-B/16 and Swin-T). This is precisely where $z$ contributes the least: on these transformer backbones, $z$ itself is a weak Textures detector (standalone far-AUROC $\approx\!74$--$78$ on Textures versus $\approx\!89$ on iNaturalist). Because DTD's pure-texture crops lack object or scene structure, they deviate little within the predicted class's subspace. Fusing this weak-far channel at a fixed $\beta{=}0.5$ therefore dilutes, rather than sharpens, the already-strong base separation on Textures. Nevertheless, overall far-OOD improves in 321 of the 378 cells, and where drops occur they are marginal, never exceeding $\approx\!1.8$ points. Full per-backbone, per-dataset values are provided in Appendix~\ref{app:fullresults}.

\begin{table}[t]
\centering
\small
\setlength{\tabcolsep}{4pt}
\resizebox{0.95\columnwidth}{!}{%
\begin{tabular}{@{}l cc ccc@{}}
\toprule
\multirow{2}{*}{\textbf{Backbone}} & \multicolumn{2}{c}{\textbf{Near-OOD}} & \multicolumn{3}{c}{\textbf{Far-OOD}} \\
\cmidrule(lr){2-3}\cmidrule(lr){4-6}
 & \multicolumn{1}{c}{\textbf{SSB-hard}} & \multicolumn{1}{c}{\textbf{NINCO}} & \multicolumn{1}{c}{\textbf{iNaturalist}} & \multicolumn{1}{c}{\textbf{Textures}} & \multicolumn{1}{c}{\textbf{OpenImage-O}} \\
\midrule
ResNet-18 & \multicolumn{1}{c}{\textbf{18}/0} & \multicolumn{1}{c}{\textbf{18}/0} & \multicolumn{1}{c}{17/1} & \multicolumn{1}{c}{13/5} & \multicolumn{1}{c}{14/4} \\
\quad{\scriptsize\itshape Max} & {\scriptsize $+10.86$} & {\scriptsize $+9.30$} & {\scriptsize $+27.68$} & {\scriptsize $+12.79$} & {\scriptsize $+15.04$} \\
\quad{\scriptsize\itshape Min} & {\scriptsize $+1.67$} & {\scriptsize $+0.17$} & {\scriptsize $-0.82$} & {\scriptsize $-1.49$} & {\scriptsize $-1.05$} \\
\midrule
ResNet-50 & \multicolumn{1}{c}{\textbf{18}/0} & \multicolumn{1}{c}{\textbf{18}/0} & \multicolumn{1}{c}{\textbf{18}/0} & \multicolumn{1}{c}{13/5} & \multicolumn{1}{c}{\textbf{18}/0} \\
\quad{\scriptsize\itshape Max} & {\scriptsize $+15.19$} & {\scriptsize $+10.93$} & {\scriptsize $+22.45$} & {\scriptsize $+6.98$} & {\scriptsize $+13.74$} \\
\quad{\scriptsize\itshape Min} & {\scriptsize $+2.47$} & {\scriptsize $+1.07$} & {\scriptsize $+0.06$} & {\scriptsize $-0.60$} & {\scriptsize $+0.15$} \\
\midrule
Wide-ResNet50-2 & \multicolumn{1}{c}{\textbf{18}/0} & \multicolumn{1}{c}{17/1} & \multicolumn{1}{c}{17/1} & \multicolumn{1}{c}{15/3} & \multicolumn{1}{c}{16/2} \\
\quad{\scriptsize\itshape Max} & {\scriptsize $+11.92$} & {\scriptsize $+11.01$} & {\scriptsize $+14.81$} & {\scriptsize $+11.43$} & {\scriptsize $+10.94$} \\
\quad{\scriptsize\itshape Min} & {\scriptsize $+1.35$} & {\scriptsize $-0.23$} & {\scriptsize $-0.02$} & {\scriptsize $-0.84$} & {\scriptsize $-0.55$} \\
\midrule
ConvNeXt-B & \multicolumn{1}{c}{\textbf{18}/0} & \multicolumn{1}{c}{\textbf{18}/0} & \multicolumn{1}{c}{\textbf{18}/0} & \multicolumn{1}{c}{16/2} & \multicolumn{1}{c}{\textbf{18}/0} \\
\quad{\scriptsize\itshape Max} & {\scriptsize $+19.84$} & {\scriptsize $+25.30$} & {\scriptsize $+44.16$} & {\scriptsize $+20.26$} & {\scriptsize $+27.32$} \\
\quad{\scriptsize\itshape Min} & {\scriptsize $+1.91$} & {\scriptsize $+1.10$} & {\scriptsize $+0.24$} & {\scriptsize $-0.16$} & {\scriptsize $+0.18$} \\
\midrule
MaxViT-T & \multicolumn{1}{c}{\textbf{18}/0} & \multicolumn{1}{c}{\textbf{18}/0} & \multicolumn{1}{c}{\textbf{18}/0} & \multicolumn{1}{c}{13/5} & \multicolumn{1}{c}{16/2} \\
\quad{\scriptsize\itshape Max} & {\scriptsize $+17.28$} & {\scriptsize $+22.73$} & {\scriptsize $+37.43$} & {\scriptsize $+20.35$} & {\scriptsize $+25.29$} \\
\quad{\scriptsize\itshape Min} & {\scriptsize $+1.25$} & {\scriptsize $+0.51$} & {\scriptsize $+0.11$} & {\scriptsize $-0.75$} & {\scriptsize $-0.40$} \\
\midrule
Swin-T & \multicolumn{1}{c}{\textbf{18}/0} & \multicolumn{1}{c}{\textbf{18}/0} & \multicolumn{1}{c}{16/2} & \multicolumn{1}{c}{9/9} & \multicolumn{1}{c}{16/2} \\
\quad{\scriptsize\itshape Max} & {\scriptsize $+22.41$} & {\scriptsize $+26.22$} & {\scriptsize $+37.82$} & {\scriptsize $+22.97$} & {\scriptsize $+29.22$} \\
\quad{\scriptsize\itshape Min} & {\scriptsize $+2.53$} & {\scriptsize $+0.80$} & {\scriptsize $-0.19$} & {\scriptsize $-1.62$} & {\scriptsize $-0.42$} \\
\midrule
ViT-B/16 & \multicolumn{1}{c}{\textbf{18}/0} & \multicolumn{1}{c}{\textbf{18}/0} & \multicolumn{1}{c}{16/2} & \multicolumn{1}{c}{9/9} & \multicolumn{1}{c}{15/3} \\
\quad{\scriptsize\itshape Max} & {\scriptsize $+18.48$} & {\scriptsize $+21.70$} & {\scriptsize $+31.64$} & {\scriptsize $+14.77$} & {\scriptsize $+23.47$} \\
\quad{\scriptsize\itshape Min} & {\scriptsize $+1.90$} & {\scriptsize $+0.38$} & {\scriptsize $-0.56$} & {\scriptsize $-1.75$} & {\scriptsize $-0.78$} \\
\bottomrule
\end{tabular}%
}
\caption{\textbf{Per-dataset breadth across seven backbones.} For each backbone and OOD dataset, the top line counts how many of the 18 baselines \emph{improve}\,/\,\emph{drop} in AUROC when PLUS is added (improved/dropped; \textbf{bold} marks a clean sweep where all 18 improve). The \emph{Max}\,/\,\emph{Min} lines give the largest\,/\,smallest per-detector AUROC change, in points, over the 18 detectors.}
\label{tab:cellcounts}
\end{table}

\subsection{Comparison to State-of-the-Art}

We evaluate an exemplary configuration, PLUS instantiated as MDS++ fused with $0.5z$, where our base-agnostic complement is applied to a strong base detector. We compare it against the two strongest baselines per split (near- and far-AUROC), as summarized in Table~\ref{tab:sota}.

\begin{table}[ht]
\centering
\resizebox{\linewidth}{!}{%
\begin{tabular}{@{} l c l c l @{}}
\toprule
\multirow{2}{*}{\textbf{Backbone}} & \multicolumn{2}{c}{\textbf{Near-OOD}} & \multicolumn{2}{c}{\textbf{Far-OOD}} \\
\cmidrule(lr){2-3}\cmidrule(lr){4-5}
 & \textbf{MDS++\,+\,PLUS} & \textbf{Best baselines} & \textbf{MDS++\,+\,PLUS} & \textbf{Best baselines} \\
\midrule
ViT-B/16  & \textbf{81.09} & RMDS++ \underline{80.07}, RMDS 79.99 & 92.26 & MDS++ \textbf{93.29}, ViM \underline{92.93} \\
ResNet-50 & \textbf{78.23} & ASH \underline{78.17}, NNGuide 77.53 & \underline{95.45} & ASH \textbf{95.74}, MDS++ 95.13 \\
Swin-T    & \textbf{80.86} & RMDS++ \underline{79.70}, GEN 78.97 & 93.00 & MDS++ \textbf{93.67}, ViM \underline{93.26} \\
\bottomrule
\end{tabular}%
}
\caption{\textbf{Comparison with state-of-the-art methods in AUROC.} In each Near- and Far-OOD block, the best result is in \textbf{bold}, and the second-best is \underline{underlined}.}

\label{tab:sota}
\end{table}

MDS++\,+\,PLUS achieves a higher near-OOD AUROC than every standalone baseline on all three backbones, exceeding even the strongest Mahalanobis variant (RMDS++) on ViT-B/16 and Swin-T, and edging the strongest baseline on ResNet-50 (ASH; $78.23$ vs.\ $78.17$). On far-OOD, performance is slightly below the strongest baseline (MDS++ on ViT-B/16 and Swin-T), reflecting the small, bounded far-OOD cost of incorporating the near-OOD signal $z$. Other bases fused with PLUS reach even higher near-OOD AUROC (e.g., ASH\,+\,PLUS $81.34$ on ResNet-50, Table~\ref{tab:headline}).

%% file: sec/6_analysis.tex
\section{Analysis}
\label{sec:analysis} 

\subsection{Where Does the Near-OOD Gain Come From?}
\label{sec:ablation_metric}

$z$ combines a per-class \emph{subspace} ($S_c$) and a per-class \emph{metric} ($\Sigma_c$). We ablate both independently (Table~\ref{tab:metric_ablation}).

\begin{table}[h]
\centering
\resizebox{\columnwidth}{!}{%
\begin{tabular}{@{}l l ccc@{}}
\toprule
\textbf{Subspace} & \textbf{Metric} & \textbf{ViT-B/16} & \textbf{ResNet-50} & \textbf{Swin-T} \\
\midrule
\multirow{3}{*}{SAE top-16 (ours)}
  & per-class $\Sigma_c$ & \textbf{81.23} & \textbf{78.22} & \textbf{80.90} \\
  & tied & 80.61 & 74.79 & 80.35 \\
  & isotropic & 80.33 & 75.46 & 80.04 \\
\cmidrule(l){2-5}
\multirow{3}{*}{Random orthonormal}
  & per-class $\Sigma_c$ & 79.57 & 74.47 & 78.49 \\
  & tied & 79.01 & 74.20 & 77.88 \\
  & isotropic & 79.00 & 74.20 & 77.88 \\
\midrule
\multicolumn{2}{@{}l}{Full-space per-class Ledoit--Wolf~\cite{ledoit2004wolf}} & 80.52 & 75.58 & 79.11 \\
\multicolumn{2}{@{}l}{MDS++ alone (no $z$)} & 79.87 & 74.71 & 78.92 \\
\bottomrule
\end{tabular}%
}
\caption{\textbf{Subspace $\times$ metric ablation} (near-OOD AUROC of MDS++\,+\,$0.5\,z$, \%). Rows vary the $16$-dim per-class subspace and the metric applied inside it. \emph{Tied}: a single within-class covariance shared by all classes, projected into each class's subspace; \emph{isotropic}: unwhitened $\lVert D\rVert^2$. Random orthonormal: mean over 5 seeds (std $\le 0.13$). Ledoit--Wolf replaces $z$ by a per-class shrinkage Mahalanobis on the full feature.}
\label{tab:metric_ablation}
\end{table}

\noindent\textbf{Neither ingredient suffices alone.} With a random $16$-dimensional subspace, the per-class metric yields \emph{no} near-OOD gain over MDS++ alone. Conversely, within the SAE subspace, a shared (tied) or isotropic metric recovers only part of the gain, and the tied metric gives almost no benefit on ResNet-50. Only their combination yields the full improvement ($+1.36$ to $+3.51$ points over MDS++). The near-OOD signal thus requires modeling per-class anisotropy along the directions in which that class is expressed.

\vspace{0.7em}

\noindent\textbf{Why an SAE supplies the subspace.} Under the same per-class metric, $16$ randomly chosen SAE neurons yield essentially no gain over MDS++ ($79.34$, $74.83$, and $78.00$ near-AUROC), so the benefit comes from selecting each class's most active neurons. Per-class PCA, the natural SAE-free alternative, is slightly ahead on ViT-B/16 and Swin-T ($81.64$ and $81.42$) but drops on ResNet-50 ($76.69$ vs.\ $78.22$), likely because it estimates each class's directions from that class's samples alone. The SAE support is thus the more consistent choice across backbones: on all seven backbones, its largest shortfall from the best subspace is $0.93$ points, versus $1.91$ for per-class PCA (Appendix~\ref{app:sae-vs-pca}).

\subsection{Sensitivity to the Fusion Weight \texorpdfstring{$\beta$}{beta}}
\label{sec:beta}
\begin{figure}[htbp]
    \centering
    \includegraphics[width=\linewidth]{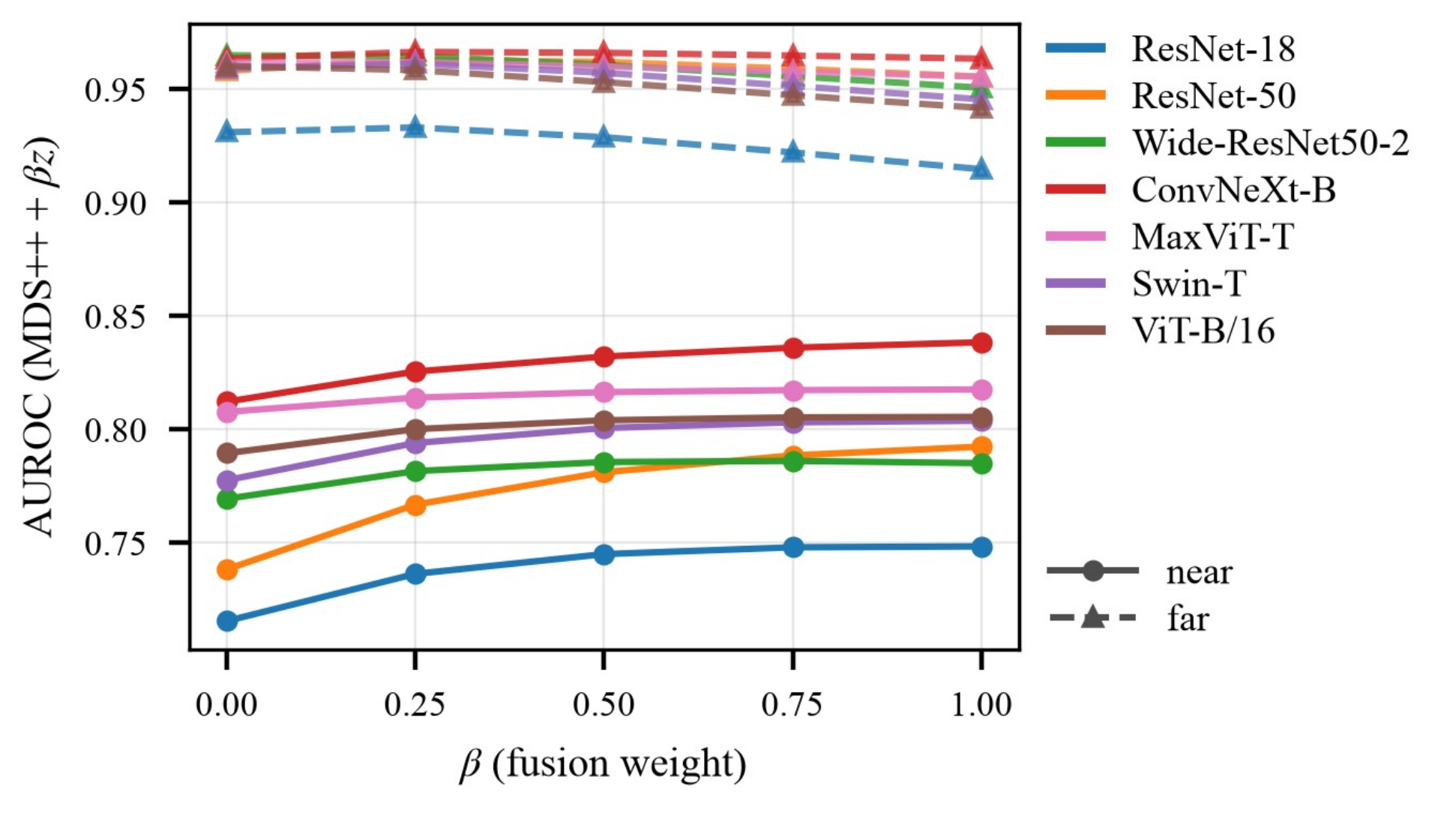}
    \caption{\textbf{Sensitivity to the fusion weight $\beta$ on an MDS++ base.} Near-AUROC (solid) and far-AUROC (dashed) across backbones.}
    \label{fig:beta}
\end{figure}

Figure~\ref{fig:beta} sweeps the fusion weight $\beta$ over an MDS++ base. Adding the $z$ complement improves near-OOD universally, but the gain profile is backbone-dependent: on ViT-B/16 and Swin-T, near-AUROC saturates around $\beta\!\approx\!0.5$, whereas on ResNet-50 and \mbox{ConvNeXt-B} it continues to rise. In contrast, far-OOD performance typically peaks near $\beta{=}0.25$ before declining. Thus, on most backbones, weights beyond $0.5$ yield little additional near-OOD benefit while steadily eroding far-OOD accuracy.

We adopt a fixed $\beta{=}0.5$ as a conservative, base-favored default (a 2:1 trust ratio). While $\beta{=}0.25$ might offer a marginally better far-OOD trade-off depending on the backbone, we exclude OOD-based tuning. This fixed value robustly secures near-gains across all seven backbones (Section~\ref{sec:cellcounts}) while avoiding the far-OOD penalties of larger weights.

\subsection{Sensitivity to the Support Size \texorpdfstring{$m$}{m}}
\label{sec:msize}

\begin{figure}[ht]
\centering
\includegraphics[width=\linewidth]{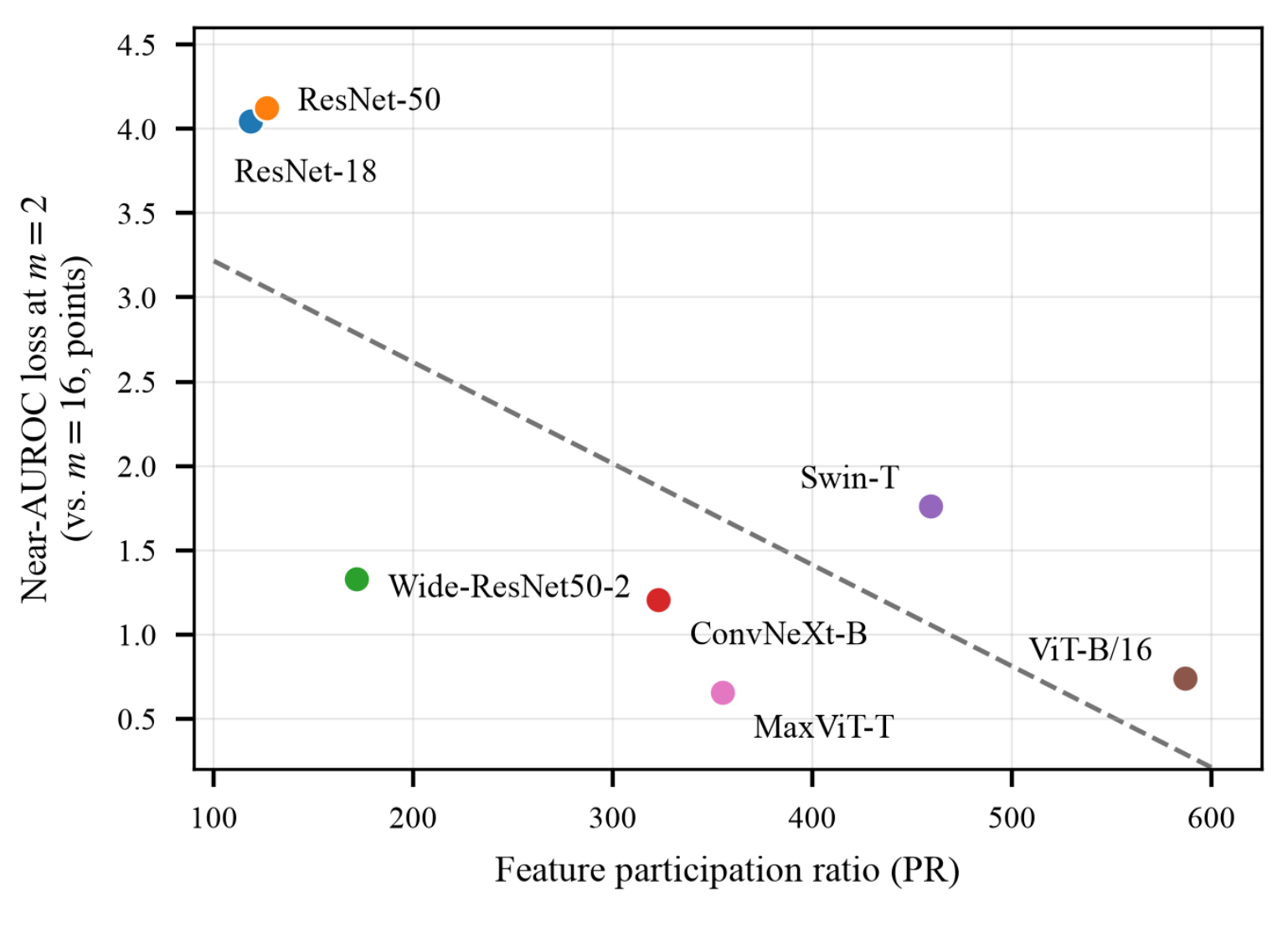}
\caption{\textbf{Support-size demand tracks the feature participation ratio.} The small-$m$ Near-AUROC loss (fused near-gain at $m{=}16$ minus $m{=}2$, in points) against the ID-train participation ratio $\mathrm{PR}$, over seven backbones.}
\label{fig:msize}
\end{figure}
The support size is fixed at $m{=}16$ for all backbones (Section~\ref{sec:experimental_setup}). This sweep examines whether a smaller support would have been preferable: across all backbones, the fused near-OOD gain increases up to $m{=}16$, and reducing the support offers no Pareto improvement. On ViT-B/16 and Swin-T, setting $m{\in}[2, 4]$ recovers a fraction of far-OOD performance only at the steep expense of near-OOD gains. On ResNet-50, performance outright collapses: $11/18$ detectors suffer far-OOD drops at $m{=}2$, compared to a clean sweep of improvements at $m{=}16$.  

This sensitivity to support size is fundamentally tied to a backbone's feature participation ratio $\mathrm{PR}=\big(\sum_i \lambda_i\big)^2\big/\sum_i \lambda_i^2$, computed over the eigenvalues $\lambda_i$ of the ID-train feature covariance ($\mathrm{PR}$ is high when variance is spread over many directions, low when a few dominate). As Figure~\ref{fig:msize} shows, the near-OOD penalty for using a small $m$ falls systematically as $\mathrm{PR}$ rises: the two lowest-$\mathrm{PR}$ backbones incur by far the largest penalty and require the full $m{=}16$, whereas the high-$\mathrm{PR}$ transformers and \mbox{ConvNeXt-B} lose little at small $m$ and accommodate $m{=}16$ without compromising near-OOD gains.

\begin{table}[h]
\centering
\small
\setlength{\tabcolsep}{3pt}
\resizebox{\columnwidth}{!}{%
\begin{tabular}{@{}l cc c cc@{}}
\toprule
 & \multicolumn{2}{c}{\textbf{Mean near gain}} & \textbf{PLUS $>$} & \multicolumn{2}{c}{\textbf{$z$ on top of $A{+}B$}} \\
\cmidrule(lr){2-3}\cmidrule(lr){5-6}
\textbf{Backbone} & $A{+}z$ & $A{+}B$ & \textbf{MDS++ partner} & \textbf{pairs improved} & \textbf{near / far $\Delta$} \\
\midrule
ViT-B/16        & $+6.48$ & $+3.88$ & 15/17 & 102/102 & $+4.22$ / $+2.15$ \\
ResNet-50       & $+4.83$ & $+1.86$ & 17/17 & 102/102 & $+3.43$ / $+1.75$ \\
Swin-T          & $+6.55$ & $+4.14$ & 14/17 & 102/102 & $+4.09$ / $+2.41$ \\
ResNet-18       & $+3.79$ & $+1.66$ & 16/17 & 102/102 & $+2.81$ / $+1.92$ \\
Wide-ResNet50-2 & $+3.12$ & $+2.92$ & 11/17 & 102/102 & $+1.71$ / $+0.90$ \\
ConvNeXt-B      & $+7.75$ & $+4.90$ & 16/17 & 102/102 & $+5.09$ / $+6.25$ \\
MaxViT-T        & $+5.84$ & $+4.22$ & \phantom{0}9/17 & 102/102 & $+3.58$ / $+2.67$ \\
\bottomrule
\end{tabular}%
}
\caption{\textbf{Ensemble control} (near-OOD AUROC). \emph{Mean near gain}: over the 18 bases $A$, the gain of $A{+}z$ (PLUS) and of $A{+}B$ averaged over the available partners $B \neq A$, both relative to $A$ alone. \emph{PLUS $>$ MDS++ partner}: number of bases ($A \neq$ MDS++) for which $A{+}z$ beats $A{+}\text{MDS++}$. \emph{$z$ on top of $A{+}B$}: adding $0.5\,c_z$ to every two-detector ensemble $c_A + 0.5\,c_B$.}
\label{tab:ensemble}
\end{table}

\subsection{Complementarity Beyond Ensembling}
\label{sec:ensemble}

Fusing two detectors often helps on its own, so we test whether $z$ adds something unique by replacing it in the fusion with an existing detector $B$ ($c_A + 0.5\,c_B$, same calibration) for six strong partners $B \in$ \{MDS++, RMDS++, ViM, KNN, GEN, ASH\} (Table~\ref{tab:ensemble}). On every backbone, adding $z$ improves a base more than adding an existing detector does on average, and it beats a strong partner, MDS++, for a majority of bases (9--17 out of 17). More importantly, adding $z$ \emph{on top of} a two-detector ensemble improves near-OOD AUROC in every one of the 714 base--partner pairs across seven backbones ($+1.7$ to $+5.1$ points on average). The information in $z$ is therefore not supplied by combining existing detectors, which is consistent with $z$ being only partially correlated with the base scores (Spearman $\rho \le 0.83$).

%% file: sec/7_conclusion.tex
\section{Conclusion}
\label{sec:conclusion}

We present PLUS, a detector-agnostic complement for near-OOD detection that computes a per-class subspace deviation $z$ from a fixed sparse autoencoder support. Fused additively with any calibrated base score under a shared protocol, the signal $z$ enables robust near-OOD discrimination. Across seven distinct backbones and 18 post-hoc detectors spanning parametric, non-parametric, logit-based, prototype, and hybrid families, adding $z$ consistently improves near-OOD AUROC for every evaluated combination while broadly preserving far-OOD performance. Crucially, these gains require no network retraining, architectural modifications, or exposure to auxiliary OOD data. This empirical improvement is consistent with our geometric characterization: while near-OOD inputs remain ID-like under global feature summaries, they exhibit measurable deviation within their predicted class's low-dimensional subspace. The subspace signal $z$ captures this within-manifold cue precisely where global scores lose discriminative power, thereby complementing far-saturated detectors at a bounded far-OOD cost. PLUS integrates this geometric insight with sparse-autoencoder-based per-class subspaces, scoring them via a unified, detector-agnostic protocol. This yields a simple, reproducible, and effective framework for reliable OOD deployment.

\vspace{0.7em}
\noindent\textbf{Future Work.} Natural next steps include generalizing per-class subspace construction beyond ImageNet to small-scale datasets as well as enterprise-scale settings with massive label spaces. Furthermore, extending this framework from image-level classification to dense prediction tasks, such as object detection, offers a promising direction. From a deployment perspective, replacing the fixed weighting with a principled, sample-adaptive fusion mechanism without auxiliary OOD exposure would further maximize practical sensitivity. Finally, establishing a rigorous mathematical framework for subspace separability will clarify the fundamental conditions under which per-class deviations yield near-OOD discriminability.

\section*{Acknowledgments}
This work was supported by the National Research Foundation of Korea grant (No. RS-2026-25490405) and by the Institute of Information \& Communications Technology Planning \& Evaluation (IITP) grant (No. RS-2025-02214652) funded by the Korea government (MSIT).

%% file: sec/99_appendix.tex
\section{Backbone Checkpoints}
\label{app:backbone}

\begin{table*}[ht]
\centering
\small
\begin{tabular}{lll r c}
\toprule
Backbone & Family & torchvision checkpoint & \(d\) & OpenOOD \\
\midrule
ResNet-18       & CNN, BasicBlock           & \texttt{ResNet18\_Weights.IMAGENET1K\_V1}          & 512  &            \\
ResNet-50       & CNN, Bottleneck           & \texttt{ResNet50\_Weights.IMAGENET1K\_V1}          & 2048 & \checkmark \\
Wide-ResNet50-2 & CNN, Bottleneck ($2\times$ width) & \texttt{Wide\_ResNet50\_2\_Weights.IMAGENET1K\_V1} & 2048 &            \\
ConvNeXt-B      & ConvNeXt                  & \texttt{ConvNeXt\_Base\_Weights.IMAGENET1K\_V1}    & 1024 &            \\
Swin-T          & Swin Transformer          & \texttt{Swin\_T\_Weights.IMAGENET1K\_V1}           & 768  & \checkmark \\
ViT-B/16        & ViT                       & \texttt{ViT\_B\_16\_Weights.IMAGENET1K\_V1}        & 768  & \checkmark \\
MaxViT-T        & Hybrid Conv-Transformer   & \texttt{MaxVit\_T\_Weights.IMAGENET1K\_V1}         & 512  &            \\
\bottomrule
\end{tabular}
\caption{\textbf{Backbone checkpoints.} All weights are torchvision \texttt{IMAGENET1K\_V1} public checkpoints, used without fine-tuning; \(d\) is the penultimate-feature (pre-logit) dimension. The three ResNet-family backbones differ structurally: ResNet-18 uses shallow BasicBlock residual units (\(d{=}512\)), whereas ResNet-50 and Wide-ResNet50-2 use the deeper Bottleneck design (\(d{=}2048\)), the latter doubling the bottleneck width. Backbones marked \checkmark{} are the three standard OpenOOD v1.5 ImageNet backbones.}
\label{tab:ckpt}
\end{table*}

\section{Neuron Count \texorpdfstring{$p$}{p}: Sensitivity and Divergence}
\label{app:psweep}
\begin{figure}[htbp]
\centering
\includegraphics[width=0.85\linewidth]{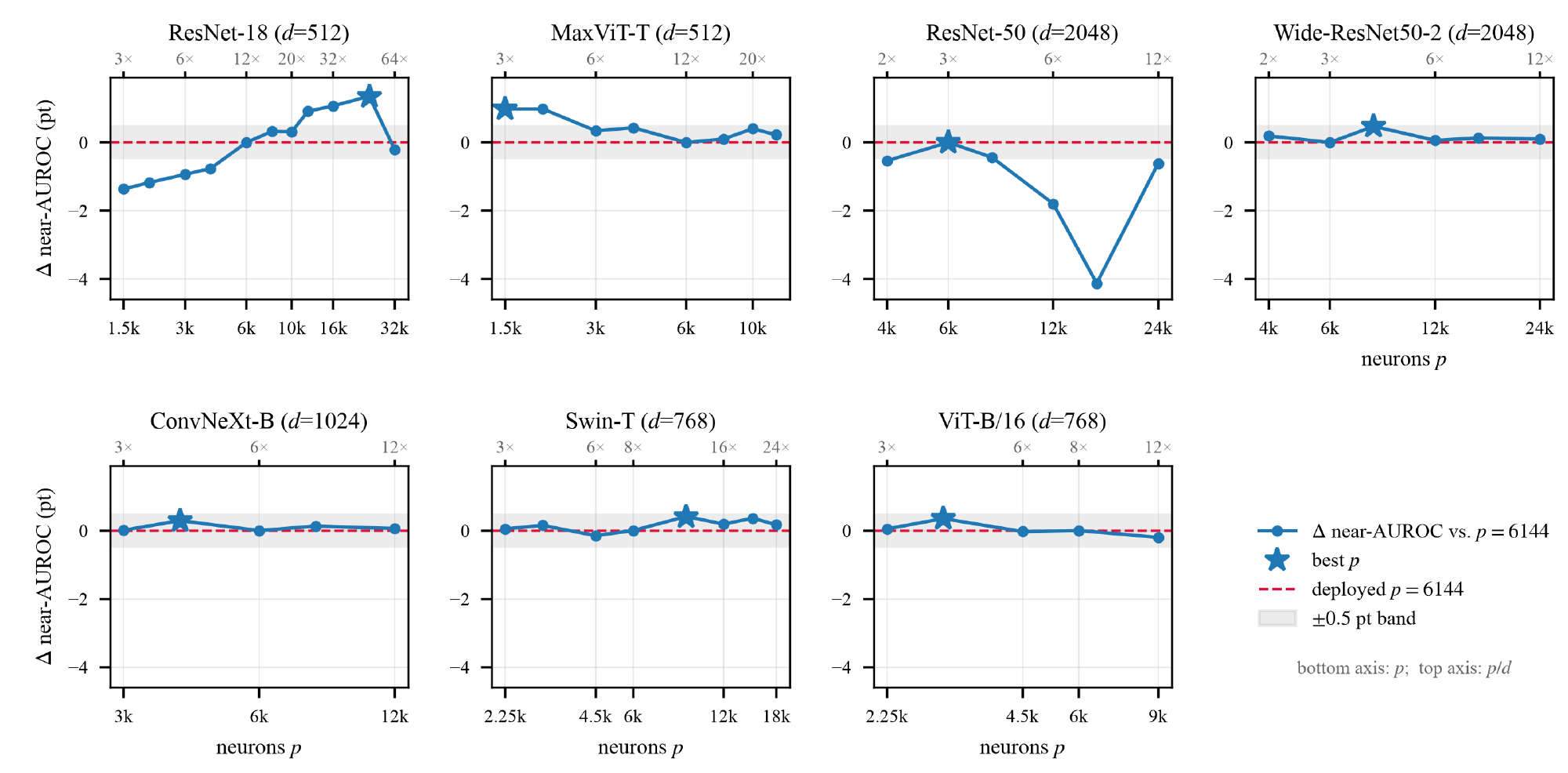}
\caption{\textbf{Near-AUROC response to the neuron count $p$} ($\Delta$ relative to the deployed $p = 6144$; shared vertical scale, grey band $= \pm 0.5$ pt). Four backbones remain inside the noise band (flat). ResNet-50 dips once at $8\times$ but peaks at the deployed $p$, while ResNet-18 (which seeks capacity) and MaxViT-T (which prefers compactness) diverge from the band in opposite directions. Top axis indicates the expansion factor $p/d$.}
\label{fig:psweep-delta}
\end{figure}

To evaluate the effect of fixing $p = 6144$, we sweep the SAE neuron count from $p = 2d$ to $p = 12d$ across seven backbones, extending it beyond $12d$ for ResNet-18 (up to $64d$, as its response keeps increasing), MaxViT-T, and Swin-T (up to $24d$). Holding other hyperparameters constant, we measure the $z$-standalone near- and far-OOD AUROC using ID-validation negatives against each OOD set. Table~\ref{tab:psweep} and Figure~\ref{fig:psweep-delta} detail each backbone's optimal capacity and the performance gaps compared to $p = 6144$.

\vspace{0.7em}
\noindent\textbf{Minimal standalone cost from fixing $p = 6144$.}
The near-OOD performance is generally stable with respect to $p$. Four of the seven backbones stay within $\pm 0.5$ AUROC points of their performance at the deployed $p$ across the entire sweep. Although ResNet-18 and MaxViT-T show directional trends, the largest performance gap from fixing $p = 6144$ within a practical overcomplete range ($\le\!12\times$) is $+0.98$ for MaxViT-T. While larger capacities ($>12\times$) can yield marginal improvements for certain backbones, such sizes are computationally inefficient.

\vspace{0.7em}
\noindent\textbf{Performance gaps further narrow under fusion.}
When evaluating the fused scores (base detector + $0.5z$) across all 126 combinations, tuning $p$ per backbone improves the near-OOD AUROC by only $+0.23$ on average (Table~\ref{tab:psweep}). Standalone gains are substantially reduced after fusion (e.g., the $+0.98$ gap for MaxViT-T decreases to $+0.19$). Furthermore, tuning $p$ specifically for near-OOD can degrade far-OOD performance, as observed in ViT-B/16. Consequently, backbone-specific tuning provides negligible fused gains at the cost of making a deployment constant dependent on OOD access.

\begin{table}[h]
\centering
\small
\resizebox{0.60\linewidth}{!}{%
\begin{tabular}{lrrrrr}
\toprule
Backbone & $d$ & Near-opt $p$ ($\times$) & $\Delta$near standalone & $\Delta$near fused & Far-opt $p$ ($\times$) \\
\midrule
ResNet-18       & 512  & 24576 ($48\times$) & $+1.35$ & $+0.75$ & 2048 ($4\times$) \\
ResNet-50       & 2048 & 6144 ($3\times$)   & $+0.00$ & $+0.00$ & 4096 ($2\times$) \\
Wide-ResNet50-2 & 2048 & 8192 ($4\times$)   & $+0.47$ & $+0.20$ & 4096 ($2\times$) \\
ConvNeXt-B      & 1024 & 4096 ($4\times$)   & $+0.29$ & $+0.12$ & 4096 ($4\times$) \\
MaxViT-T        & 512  & 1536 ($3\times$)   & $+0.98$ & $+0.19$ & 3072 ($6\times$) \\
Swin-T          & 768  & 9216 ($12\times$)  & $+0.41$ & $+0.20$ & 15360 ($20\times$) \\
ViT-B/16        & 768  & 3072 ($4\times$)   & $+0.35$ & $+0.15$ & 6144 ($8\times$) \\
\bottomrule
\end{tabular}%
}
\caption{\textbf{Per-backbone optimal SAE capacity.} Near-opt and Far-opt $p$ indicate the capacities maximizing standalone near- and far-OOD AUROC, respectively. $\Delta$near std and fused represent the near-AUROC gained by using the near-optimal $p$ instead of the fixed $p=6144$, evaluated as $z$ standalone (standalone) and after base+$0.5z$ fusion (fused).}
\label{tab:psweep}
\end{table}

\begin{figure}[htbp]
\centering
\includegraphics[width=0.85\linewidth]{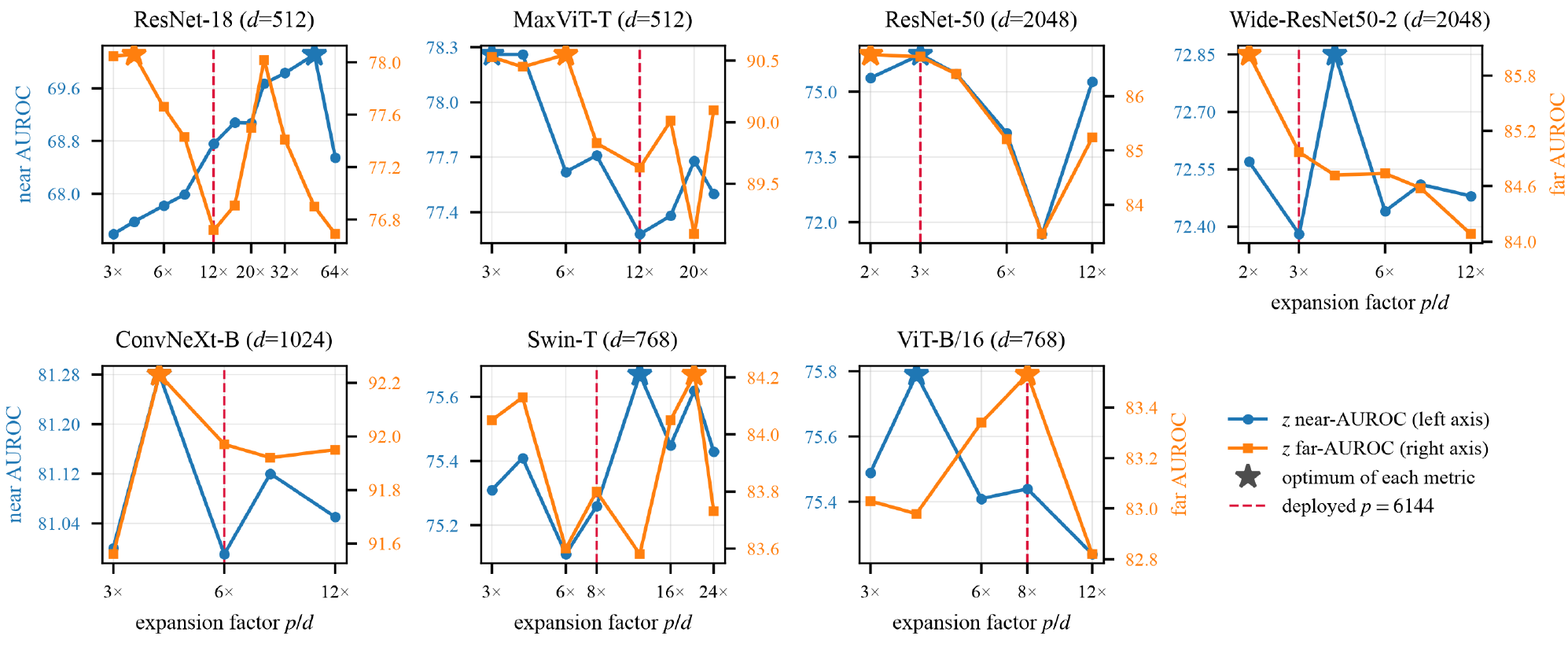}
\caption{\textbf{Near (blue) and far (orange) $z$-AUROC vs.\ neuron count.} Stars mark each metric's optimum, while the red dashed line indicates the deployed $p = 6144$. The near- and far-optima diverge for six of the seven backbones; CNNs prefer a smaller $p$ for far-OOD, whereas transformers prefer a larger one.}
\label{fig:psweep-nearfar}
\end{figure}

\vspace{0.7em}
\noindent\textbf{Divergent optima for near- and far-OOD.}
For six of the seven backbones, the optimal $p$ for near-OOD differs from that for far-OOD (Table~\ref{tab:psweep}, Fig.~\ref{fig:psweep-nearfar}). This divergence follows architectural lines: standard CNNs favor a smaller $p$ for far-OOD detection, whereas transformers and hybrids benefit from a larger $p$. Increasing $p$ allows the per-class support to capture finer ID-specific directions, which improves near-OOD separation but can overfit and degrade far-OOD performance (e.g., ResNet-18 is near-optimal at $48\times$ but far-optimal at $4\times$). ConvNeXt-B is the sole exception where these optima align. Ultimately, the fixed capacity of $p = 6144$ serves as an effective compromise between these competing optima.

\vspace{0.7em}
\noindent\textbf{Summary.}
We fix $p = 6144$ across all backbones. The standalone performance concession is at most $0.98$ near-AUROC points within $12\times$ ($1.35$ for ResNet-18 at $48\times$), and this gap shrinks to $+0.23$ on average after fusion.

\section{Full Experimental Results}
\label{app:fullresults}

For each backbone, all 18 baselines are reported both standalone and with PLUS (shaded, \textbf{bold}), under the same protocol as Table~\ref{tab:headline}. An \underline{underline} marks a per-metric performance regression.

\begin{table}[htbp]
\centering
\scriptsize
\setlength{\tabcolsep}{2.5pt}
\renewcommand{\arraystretch}{1.12}
\resizebox{0.8\textwidth}{!}{%
\begin{tabular}{@{}l *{14}{c}@{}}
\toprule
\multirow{2}{*}{\textbf{Method}} & \multicolumn{2}{c}{\textbf{SSB-hard}} & \multicolumn{2}{c}{\textbf{NINCO}} & \multicolumn{2}{c}{\textbf{Near-OOD Average}} & \multicolumn{2}{c}{\textbf{iNaturalist}} & \multicolumn{2}{c}{\textbf{Textures}} & \multicolumn{2}{c}{\textbf{OpenImage-O}} & \multicolumn{2}{c}{\textbf{Far-OOD Average}} \\
\cmidrule(lr){2-3}\cmidrule(lr){4-5}\cmidrule(lr){6-7}\cmidrule(lr){8-9}\cmidrule(lr){10-11}\cmidrule(lr){12-13}\cmidrule(lr){14-15}
 & \textbf{AUROC}$\uparrow$ & \textbf{FPR95}$\downarrow$ & \textbf{AUROC}$\uparrow$ & \textbf{FPR95}$\downarrow$ & \textbf{AUROC}$\uparrow$ & \textbf{FPR95}$\downarrow$ & \textbf{AUROC}$\uparrow$ & \textbf{FPR95}$\downarrow$ & \textbf{AUROC}$\uparrow$ & \textbf{FPR95}$\downarrow$ & \textbf{AUROC}$\uparrow$ & \textbf{FPR95}$\downarrow$ & \textbf{AUROC}$\uparrow$ & \textbf{FPR95}$\downarrow$ \\
\midrule
MSP & 68.94 & 86.42 & 78.11 & 77.27 & 73.53 & 81.84 & 88.19 & 42.41 & 85.06 & 56.46 & 84.86 & 56.16 & 86.04 & 51.68 \\
\rowcolor{gray!9}\textbf{MSP\,$+$\,PLUS} & \textbf{73.66} & \textbf{79.37} & \textbf{82.60} & \textbf{58.22} & \textbf{78.13} & \textbf{68.80} & \textbf{91.70} & \textbf{35.01} & \textbf{85.74} & \textbf{52.74} & \textbf{87.88} & \textbf{43.83} & \textbf{88.44} & \textbf{43.86} \\
MaxLogit & 64.20 & 91.52 & 72.40 & 92.94 & 68.30 & 92.23 & 85.29 & 72.94 & 83.74 & 78.94 & 81.60 & 85.82 & 83.54 & 79.23 \\
\rowcolor{gray!9}\textbf{MaxLogit\,$+$\,PLUS} & \textbf{71.73} & \textbf{85.92} & \textbf{79.57} & \textbf{82.84} & \textbf{75.65} & \textbf{84.38} & \textbf{90.57} & \textbf{43.74} & \textbf{85.40} & \textbf{64.51} & \textbf{86.35} & \textbf{61.96} & \textbf{87.44} & \textbf{56.74} \\
Energy & 58.80 & 92.24 & 66.02 & 94.13 & 62.41 & 93.19 & 79.30 & 83.56 & 81.17 & 83.66 & 76.48 & 88.82 & 78.98 & 85.35 \\
\bottomrule
\end{tabular}%
}
\caption{\textbf{ViT-B/16} --- per-dataset AUROC / FPR95, standalone vs.\ PLUS.}
\end{table}

\begin{table}[htbp]
\centering
\scriptsize
\setlength{\tabcolsep}{2.5pt}
\renewcommand{\arraystretch}{1.12}
\resizebox{0.8\textwidth}{!}{%
\begin{tabular}{@{}l *{14}{c}@{}}
\toprule
\multirow{2}{*}{\textbf{Method}} & \multicolumn{2}{c}{\textbf{SSB-hard}} & \multicolumn{2}{c}{\textbf{NINCO}} & \multicolumn{2}{c}{\textbf{Near-OOD Average}} & \multicolumn{2}{c}{\textbf{iNaturalist}} & \multicolumn{2}{c}{\textbf{Textures}} & \multicolumn{2}{c}{\textbf{OpenImage-O}} & \multicolumn{2}{c}{\textbf{Far-OOD Average}} \\
\cmidrule(lr){2-3}\cmidrule(lr){4-5}\cmidrule(lr){6-7}\cmidrule(lr){8-9}\cmidrule(lr){10-11}\cmidrule(lr){12-13}\cmidrule(lr){14-15}
 & \textbf{AUROC}$\uparrow$ & \textbf{FPR95}$\downarrow$ & \textbf{AUROC}$\uparrow$ & \textbf{FPR95}$\downarrow$ & \textbf{AUROC}$\uparrow$ & \textbf{FPR95}$\downarrow$ & \textbf{AUROC}$\uparrow$ & \textbf{FPR95}$\downarrow$ & \textbf{AUROC}$\uparrow$ & \textbf{FPR95}$\downarrow$ & \textbf{AUROC}$\uparrow$ & \textbf{FPR95}$\downarrow$ & \textbf{AUROC}$\uparrow$ & \textbf{FPR95}$\downarrow$ \\
\midrule
MSP & 72.09 & 74.48 & 79.95 & 56.76 & 76.02 & 65.62 & 88.41 & 43.34 & 82.43 & 60.88 & 84.86 & 50.13 & 85.23 & 51.45 \\
\rowcolor{gray!9}\textbf{MSP\,$+$\,PLUS} & \textbf{75.98} & \textbf{71.61} & \textbf{82.61} & \textbf{55.04} & \textbf{79.30} & \textbf{63.33} & \textbf{92.69} & \textbf{33.58} & \textbf{87.62} & \textbf{46.06} & \textbf{88.62} & \textbf{43.01} & \textbf{89.64} & \textbf{40.88} \\
MaxLogit & 72.51 & 76.19 & 80.41 & 59.41 & 76.46 & 67.80 & 91.16 & 30.62 & 88.39 & 46.11 & 89.17 & 37.86 & 89.57 & 38.20 \\
\rowcolor{gray!9}\textbf{MaxLogit\,$+$\,PLUS} & \textbf{77.97} & \textbf{71.35} & \textbf{83.75} & \textbf{55.00} & \textbf{80.86} & \textbf{63.17} & \textbf{95.24} & \textbf{21.10} & \textbf{91.74} & \textbf{34.08} & \textbf{91.95} & \textbf{32.43} & \textbf{92.98} & \textbf{29.20} \\
Energy & 72.08 & 76.54 & 79.70 & 60.57 & 75.89 & 68.56 & 90.63 & 31.34 & 88.70 & 45.76 & 89.06 & 38.08 & 89.46 & 38.39 \\
\bottomrule
\end{tabular}%
}
\caption{\textbf{ResNet-50} --- per-dataset AUROC / FPR95, standalone vs.\ PLUS.}
\end{table}

\begin{table}[ht]
\centering
\scriptsize
\setlength{\tabcolsep}{2.5pt}
\renewcommand{\arraystretch}{1.12}
\resizebox{0.8\textwidth}{!}{%
\begin{tabular}{@{}l *{14}{c}@{}}
\toprule
\multirow{2}{*}{\textbf{Method}} & \multicolumn{2}{c}{\textbf{SSB-hard}} & \multicolumn{2}{c}{\textbf{NINCO}} & \multicolumn{2}{c}{\textbf{Near-OOD Average}} & \multicolumn{2}{c}{\textbf{iNaturalist}} & \multicolumn{2}{c}{\textbf{Textures}} & \multicolumn{2}{c}{\textbf{OpenImage-O}} & \multicolumn{2}{c}{\textbf{Far-OOD Average}} \\
\cmidrule(lr){2-3}\cmidrule(lr){4-5}\cmidrule(lr){6-7}\cmidrule(lr){8-9}\cmidrule(lr){10-11}\cmidrule(lr){12-13}\cmidrule(lr){14-15}
 & \textbf{AUROC}$\uparrow$ & \textbf{FPR95}$\downarrow$ & \textbf{AUROC}$\uparrow$ & \textbf{FPR95}$\downarrow$ & \textbf{AUROC}$\uparrow$ & \textbf{FPR95}$\downarrow$ & \textbf{AUROC}$\uparrow$ & \textbf{FPR95}$\downarrow$ & \textbf{AUROC}$\uparrow$ & \textbf{FPR95}$\downarrow$ & \textbf{AUROC}$\uparrow$ & \textbf{FPR95}$\downarrow$ & \textbf{AUROC}$\uparrow$ & \textbf{FPR95}$\downarrow$ \\
\midrule
MSP & 71.78 & 80.90 & 81.72 & 61.20 & 76.75 & 71.05 & 89.86 & 37.31 & 83.27 & 60.88 & 85.77 & 49.30 & 86.30 & 49.16 \\
\rowcolor{gray!9}\textbf{MSP\,$+$\,PLUS} & \textbf{75.49} & \textbf{75.00} & \textbf{84.55} & \textbf{53.65} & \textbf{80.02} & \textbf{64.33} & \textbf{92.96} & \textbf{31.73} & \textbf{84.31} & \textbf{53.85} & \textbf{88.36} & \textbf{43.31} & \textbf{88.54} & \textbf{42.96} \\
MaxLogit & 70.44 & 86.52 & 80.89 & 74.94 & 75.66 & 80.73 & 88.99 & 49.20 & 81.68 & 79.82 & 83.72 & 74.43 & 84.80 & 67.82 \\
\rowcolor{gray!9}\textbf{MaxLogit\,$+$\,PLUS} & \textbf{75.21} & \textbf{81.67} & \textbf{84.37} & \textbf{62.32} & \textbf{79.79} & \textbf{72.00} & \textbf{92.82} & \textbf{31.40} & \textbf{83.31} & \textbf{68.55} & \textbf{87.24} & \textbf{54.52} & \textbf{87.79} & \textbf{51.49} \\
Energy & 68.21 & 87.45 & 78.25 & 79.03 & 73.23 & 83.24 & 85.13 & 61.52 & 78.96 & 84.45 & 79.86 & 80.79 & 81.32 & 75.59 \\
\bottomrule
\end{tabular}%
}
\caption{\textbf{Swin-T} --- per-dataset AUROC / FPR95, standalone vs.\ PLUS.}
\end{table}

\begin{table}[ht]
\centering
\scriptsize
\setlength{\tabcolsep}{2.5pt}
\renewcommand{\arraystretch}{1.12}
\resizebox{0.8\textwidth}{!}{%
\begin{tabular}{@{}l *{14}{c}@{}}
\toprule
\multirow{2}{*}{\textbf{Method}} & \multicolumn{2}{c}{\textbf{SSB-hard}} & \multicolumn{2}{c}{\textbf{NINCO}} & \multicolumn{2}{c}{\textbf{Near-OOD Average}} & \multicolumn{2}{c}{\textbf{iNaturalist}} & \multicolumn{2}{c}{\textbf{Textures}} & \multicolumn{2}{c}{\textbf{OpenImage-O}} & \multicolumn{2}{c}{\textbf{Far-OOD Average}} \\
\cmidrule(lr){2-3}\cmidrule(lr){4-5}\cmidrule(lr){6-7}\cmidrule(lr){8-9}\cmidrule(lr){10-11}\cmidrule(lr){12-13}\cmidrule(lr){14-15}
 & \textbf{AUROC}$\uparrow$ & \textbf{FPR95}$\downarrow$ & \textbf{AUROC}$\uparrow$ & \textbf{FPR95}$\downarrow$ & \textbf{AUROC}$\uparrow$ & \textbf{FPR95}$\downarrow$ & \textbf{AUROC}$\uparrow$ & \textbf{FPR95}$\downarrow$ & \textbf{AUROC}$\uparrow$ & \textbf{FPR95}$\downarrow$ & \textbf{AUROC}$\uparrow$ & \textbf{FPR95}$\downarrow$ & \textbf{AUROC}$\uparrow$ & \textbf{FPR95}$\downarrow$ \\
\midrule
MSP & 67.37 & 78.86 & 77.49 & 60.89 & 72.43 & 69.88 & 87.07 & 45.88 & 80.45 & 63.02 & 81.47 & 56.50 & 83.00 & 55.13 \\
\rowcolor{gray!9}\textbf{MSP\,$+$\,PLUS} & \textbf{70.67} & \underline{79.18} & \textbf{79.12} & \underline{61.87} & \textbf{74.90} & \underline{70.53} & \textbf{90.38} & \textbf{41.16} & \textbf{84.93} & \textbf{53.25} & \textbf{84.20} & \textbf{53.87} & \textbf{86.50} & \textbf{49.43} \\
MaxLogit & 65.46 & 83.58 & 76.50 & 66.46 & 70.98 & 75.02 & 90.29 & 32.00 & 87.69 & 46.96 & 86.32 & 43.40 & 88.10 & 40.79 \\
\rowcolor{gray!9}\textbf{MaxLogit\,$+$\,PLUS} & \textbf{70.65} & \textbf{81.00} & \textbf{78.59} & \textbf{64.38} & \textbf{74.62} & \textbf{72.69} & \textbf{93.03} & \textbf{28.80} & \textbf{90.15} & \textbf{39.36} & \textbf{88.10} & \textbf{41.69} & \textbf{90.43} & \textbf{36.62} \\
Energy & 64.82 & 83.94 & 75.60 & 67.34 & 70.21 & 75.64 & 89.84 & 32.35 & 88.38 & 46.90 & 86.37 & 43.60 & 88.20 & 40.95 \\
\bottomrule
\end{tabular}%
}
\caption{\textbf{ResNet-18} --- per-dataset AUROC / FPR95, standalone vs.\ PLUS.}
\end{table}

\begin{table}[ht]
\centering
\scriptsize
\setlength{\tabcolsep}{2.5pt}
\renewcommand{\arraystretch}{1.12}
\resizebox{0.8\textwidth}{!}{%
\begin{tabular}{@{}l *{14}{c}@{}}
\toprule
\multirow{2}{*}{\textbf{Method}} & \multicolumn{2}{c}{\textbf{SSB-hard}} & \multicolumn{2}{c}{\textbf{NINCO}} & \multicolumn{2}{c}{\textbf{Near-OOD Average}} & \multicolumn{2}{c}{\textbf{iNaturalist}} & \multicolumn{2}{c}{\textbf{Textures}} & \multicolumn{2}{c}{\textbf{OpenImage-O}} & \multicolumn{2}{c}{\textbf{Far-OOD Average}} \\
\cmidrule(lr){2-3}\cmidrule(lr){4-5}\cmidrule(lr){6-7}\cmidrule(lr){8-9}\cmidrule(lr){10-11}\cmidrule(lr){12-13}\cmidrule(lr){14-15}
 & \textbf{AUROC}$\uparrow$ & \textbf{FPR95}$\downarrow$ & \textbf{AUROC}$\uparrow$ & \textbf{FPR95}$\downarrow$ & \textbf{AUROC}$\uparrow$ & \textbf{FPR95}$\downarrow$ & \textbf{AUROC}$\uparrow$ & \textbf{FPR95}$\downarrow$ & \textbf{AUROC}$\uparrow$ & \textbf{FPR95}$\downarrow$ & \textbf{AUROC}$\uparrow$ & \textbf{FPR95}$\downarrow$ & \textbf{AUROC}$\uparrow$ & \textbf{FPR95}$\downarrow$ \\
\midrule
MSP & 74.08 & 73.95 & 82.04 & 56.71 & 78.06 & 65.33 & 89.95 & 39.03 & 82.87 & 58.67 & 87.02 & 45.55 & 86.61 & 47.75 \\
\rowcolor{gray!9}\textbf{MSP\,$+$\,PLUS} & \textbf{75.57} & \underline{76.54} & \textbf{83.41} & \underline{57.44} & \textbf{79.49} & \underline{66.99} & \textbf{92.39} & \textbf{33.88} & \textbf{86.28} & \textbf{48.95} & \textbf{88.88} & \textbf{41.79} & \textbf{89.18} & \textbf{41.54} \\
MaxLogit & 74.97 & 76.27 & 82.41 & 59.49 & 78.69 & 67.88 & 91.27 & 31.10 & 86.04 & 54.77 & 89.75 & 37.74 & 89.02 & 41.20 \\
\rowcolor{gray!9}\textbf{MaxLogit\,$+$\,PLUS} & \textbf{77.67} & \textbf{74.00} & \textbf{84.50} & \textbf{53.65} & \textbf{81.09} & \textbf{63.83} & \textbf{94.30} & \textbf{24.23} & \textbf{89.27} & \textbf{41.83} & \textbf{91.26} & \textbf{33.81} & \textbf{91.61} & \textbf{33.29} \\
Energy & 74.69 & 76.46 & 81.76 & 59.76 & 78.22 & 68.11 & 90.39 & 31.46 & 85.89 & 54.76 & 89.42 & 38.56 & 88.57 & 41.59 \\
\bottomrule
\end{tabular}%
}
\caption{\textbf{Wide-ResNet50-2} --- per-dataset AUROC / FPR95, standalone vs.\ PLUS.}
\end{table}

\begin{table}[ht]
\centering
\scriptsize
\setlength{\tabcolsep}{2.5pt}
\renewcommand{\arraystretch}{1.12}
\resizebox{0.8\textwidth}{!}{%
\begin{tabular}{@{}l *{14}{c}@{}}
\toprule
\multirow{2}{*}{\textbf{Method}} & \multicolumn{2}{c}{\textbf{SSB-hard}} & \multicolumn{2}{c}{\textbf{NINCO}} & \multicolumn{2}{c}{\textbf{Near-OOD Average}} & \multicolumn{2}{c}{\textbf{iNaturalist}} & \multicolumn{2}{c}{\textbf{Textures}} & \multicolumn{2}{c}{\textbf{OpenImage-O}} & \multicolumn{2}{c}{\textbf{Far-OOD Average}} \\
\cmidrule(lr){2-3}\cmidrule(lr){4-5}\cmidrule(lr){6-7}\cmidrule(lr){8-9}\cmidrule(lr){10-11}\cmidrule(lr){12-13}\cmidrule(lr){14-15}
 & \textbf{AUROC}$\uparrow$ & \textbf{FPR95}$\downarrow$ & \textbf{AUROC}$\uparrow$ & \textbf{FPR95}$\downarrow$ & \textbf{AUROC}$\uparrow$ & \textbf{FPR95}$\downarrow$ & \textbf{AUROC}$\uparrow$ & \textbf{FPR95}$\downarrow$ & \textbf{AUROC}$\uparrow$ & \textbf{FPR95}$\downarrow$ & \textbf{AUROC}$\uparrow$ & \textbf{FPR95}$\downarrow$ & \textbf{AUROC}$\uparrow$ & \textbf{FPR95}$\downarrow$ \\
\midrule
MSP & 75.47 & 76.75 & 81.08 & 71.43 & 78.28 & 74.09 & 89.54 & 43.46 & 74.36 & 96.46 & 83.10 & 74.66 & 82.33 & 71.53 \\
\rowcolor{gray!9}\textbf{MSP\,$+$\,PLUS} & \textbf{79.39} & \textbf{68.19} & \textbf{85.81} & \textbf{52.05} & \textbf{82.60} & \textbf{60.12} & \textbf{93.89} & \textbf{28.11} & \textbf{81.08} & \textbf{82.38} & \textbf{88.66} & \textbf{45.27} & \textbf{87.88} & \textbf{51.92} \\
MaxLogit & 72.93 & 85.60 & 76.43 & 86.40 & 74.68 & 86.00 & 82.63 & 77.82 & 64.66 & 98.11 & 73.40 & 93.53 & 73.56 & 89.82 \\
\rowcolor{gray!9}\textbf{MaxLogit\,$+$\,PLUS} & \textbf{78.49} & \textbf{75.97} & \textbf{83.96} & \textbf{66.65} & \textbf{81.22} & \textbf{71.31} & \textbf{91.67} & \textbf{40.36} & \textbf{74.91} & \textbf{95.74} & \textbf{83.87} & \textbf{75.39} & \textbf{83.48} & \textbf{70.50} \\
Energy & 65.59 & 90.04 & 64.05 & 92.18 & 64.82 & 91.11 & 55.75 & 92.24 & 49.48 & 98.27 & 52.68 & 96.42 & 52.64 & 95.64 \\
\bottomrule
\end{tabular}%
}
\caption{\textbf{ConvNeXt-B} --- per-dataset AUROC / FPR95, standalone vs.\ PLUS.}
\end{table}

\begin{table}[ht]
\centering
\scriptsize
\setlength{\tabcolsep}{2.5pt}
\renewcommand{\arraystretch}{1.12}
\resizebox{0.8\textwidth}{!}{%
\begin{tabular}{@{}l *{14}{c}@{}}
\toprule
\multirow{2}{*}{\textbf{Method}} & \multicolumn{2}{c}{\textbf{SSB-hard}} & \multicolumn{2}{c}{\textbf{NINCO}} & \multicolumn{2}{c}{\textbf{Near-OOD Average}} & \multicolumn{2}{c}{\textbf{iNaturalist}} & \multicolumn{2}{c}{\textbf{Textures}} & \multicolumn{2}{c}{\textbf{OpenImage-O}} & \multicolumn{2}{c}{\textbf{Far-OOD Average}} \\
\cmidrule(lr){2-3}\cmidrule(lr){4-5}\cmidrule(lr){6-7}\cmidrule(lr){8-9}\cmidrule(lr){10-11}\cmidrule(lr){12-13}\cmidrule(lr){14-15}
 & \textbf{AUROC}$\uparrow$ & \textbf{FPR95}$\downarrow$ & \textbf{AUROC}$\uparrow$ & \textbf{FPR95}$\downarrow$ & \textbf{AUROC}$\uparrow$ & \textbf{FPR95}$\downarrow$ & \textbf{AUROC}$\uparrow$ & \textbf{FPR95}$\downarrow$ & \textbf{AUROC}$\uparrow$ & \textbf{FPR95}$\downarrow$ & \textbf{AUROC}$\uparrow$ & \textbf{FPR95}$\downarrow$ & \textbf{AUROC}$\uparrow$ & \textbf{FPR95}$\downarrow$ \\
\midrule
MSP & 70.99 & 84.39 & 80.86 & 64.09 & 75.92 & 74.24 & 88.76 & 40.06 & 85.51 & 54.02 & 85.85 & 49.84 & 86.71 & 47.97 \\
\rowcolor{gray!9}\textbf{MSP\,$+$\,PLUS} & \textbf{74.81} & \textbf{79.02} & \textbf{84.19} & \textbf{52.96} & \textbf{79.50} & \textbf{65.99} & \textbf{92.37} & \textbf{31.75} & \textbf{87.12} & \textbf{49.41} & \textbf{88.69} & \textbf{40.66} & \textbf{89.39} & \textbf{40.61} \\
MaxLogit & 68.16 & 87.17 & 75.89 & 85.96 & 72.03 & 86.56 & 90.06 & 47.00 & 85.87 & 69.98 & 82.06 & 81.01 & 86.00 & 66.00 \\
\rowcolor{gray!9}\textbf{MaxLogit\,$+$\,PLUS} & \textbf{73.21} & \textbf{83.52} & \textbf{81.57} & \textbf{74.80} & \textbf{77.39} & \textbf{79.16} & \textbf{93.51} & \textbf{31.39} & \textbf{87.62} & \textbf{61.35} & \textbf{86.59} & \textbf{64.50} & \textbf{89.24} & \textbf{52.41} \\
Energy & 63.88 & 88.85 & 68.31 & 91.32 & 66.09 & 90.08 & 88.16 & 55.86 & 83.65 & 76.52 & 75.68 & 88.92 & 82.50 & 73.77 \\
\bottomrule
\end{tabular}%
}
\caption{\textbf{MaxViT-T} --- per-dataset AUROC / FPR95, standalone vs.\ PLUS.}
\end{table}

\clearpage

\section{Why Construct the Subspace from SAE Neurons?}
\label{app:sae-vs-pca}

Our per-class subspace is spanned by each class's top-active SAE neurons (Section~\ref{sec:method_subspace}). Table~\ref{tab:subspace_choice} extends the comparison of Section~\ref{sec:ablation_metric} to all seven backbones: random subspaces yield essentially no gain over MDS++, and all SAE-free alternatives suffer larger worst-case shortfalls. Beyond this consistency, the SAE provides an analytical advantage over per-class PCA, detailed below.

\begin{table*}[h]
\centering
\resizebox{\textwidth}{!}{%
\begin{tabular}{@{}l ccc cccc cc@{}}
\toprule
\textbf{Subspace ($m{=}16$)} & ViT-B/16 & ResNet-50 & Swin-T & ResNet-18 & Wide-ResNet50-2 & ConvNeXt-B & MaxViT-T & Mean & Max gap \\
\midrule
MDS++ alone (no $z$) & 79.87 & 74.71 & 78.92 & 70.94 & 77.81 & 81.49 & 80.96 & 77.81 & -- \\
\rowcolor{gray!9}SAE top-active (ours) & 81.23 & \textbf{78.22} & 80.90 & 73.30 & 79.03 & 83.53 & 81.90 & \textbf{79.73} & \textbf{0.93} \\
Per-class PCA & \textbf{81.64} & 76.69 & \textbf{81.42} & 71.96 & \textbf{79.96} & \textbf{83.73} & \textbf{82.48} & 79.70 & 1.91 \\
Global PCA + per-class $\Sigma_c$ & 80.05 & 77.01 & 79.41 & \textbf{73.87} & 79.77 & 82.86 & 81.40 & 79.20 & 2.01 \\
Top-16 raw channels & 79.82 & 76.75 & 78.76 & 72.28 & 78.69 & 81.90 & 81.23 & 78.49 & 2.66 \\
Random SAE neurons$^\dagger$ & 79.34 & 74.83 & 78.00 & 71.37 & 77.59 & 81.42 & 81.05 & 77.66 & 3.42 \\
Random orthonormal$^\dagger$ & 79.57 & 74.47 & 78.49 & 71.16 & 77.59 & 81.49 & 81.10 & 77.70 & 3.75 \\
\bottomrule
\end{tabular}%
}
\caption{\textbf{Choice of the $16$-dimensional per-class subspace on all seven backbones} (near-OOD AUROC of MDS++\,+\,$0.5\,z$, \%). Every row uses the per-class metric $\Sigma_c$ and differs only in the subspace. \emph{Global PCA}: one $16$-dimensional PCA basis shared by all classes; \emph{top-16 raw channels}: each class's $16$ feature coordinates with the largest mean value. \emph{Max gap}: the largest shortfall, over the seven backbones, from the best subspace on that backbone (lower is better). Bold: best among the subspaces. $^\dagger$Mean over 5 seeds (std $\le 0.15$). On far-OOD, all non-random subspaces stay within $0.82$ points of the best on every backbone. All rows are re-estimated in the pipeline of Table~\ref{tab:metric_ablation}.}
\label{tab:subspace_choice}
\end{table*}

\vspace{0.6em}
\noindent\textbf{The SAE top-active neurons provide consistent stability.}
No single subspace is best on every backbone (Table~\ref{tab:subspace_choice}): per-class PCA leads on five of the seven backbones, global PCA on ResNet-18, and the SAE top-active neurons on ResNet-50. The SAE subspace, however, never falls far behind: its largest gap to the best subspace on any backbone is $0.93$ fused near-AUROC points (Wide-ResNet50-2), versus $1.91$ for per-class PCA, $2.01$ for global PCA, and $2.66$ for raw channels, while its seven-backbone mean is on par with per-class PCA ($79.73$ vs.\ $79.70$). Thus, while the subspace spanned by the SAE top-active neurons is the best $16$-dimensional subspace on only one of the seven backbones, it avoids the sharp near-OOD drops that every alternative incurs on some backbone. For a detector-agnostic signal designed to complement diverse backbones, we value this consistency over the backbone-specific gains it forgoes.

\vspace{0.6em}
\noindent\textbf{A shared basis for cross-class analysis.}
Each class's subspace is spanned by its top-active neurons, which represent the most characteristic directions of that class, selected from the global dictionary of a shared SAE. Because every class selects from the same dictionary, the characteristic neurons of different classes overlap explicitly and traceably; per-class PCA instead assigns each class a distinct orthonormal basis with no shared elements, precluding cross-class correspondence. This overlap is semantically structured: related classes reuse the same characteristic neurons substantially more than unrelated ones (Table~\ref{tab:sharing}). Across the three main backbones, the $118$ ImageNet dog breeds reuse the same characteristic neurons at $2.7$--$8.9\times$ the rate of a size-matched random class group, and each class shares neurons with its nearest neighbor by feature centroid at $9$--$85\times$ the random-pair rate. The SAE top-active neurons thus provide shared axes that recur across related classes, establishing a basis for cross-class analysis that a per-class PCA basis does not afford.

\begin{table}[htbp]
\centering\small
\setlength{\tabcolsep}{5pt}
\begin{tabular}{@{}lccccc@{}}
\toprule
\textbf{Backbone} & Random pair & Nearest class & Dogs (118) & Snakes (17) & Birds (18) \\
\midrule
ResNet-50 & 0.032 & 0.296\,($9\times$) & 0.087\,($2.7\times$) & 0.154\,($4.6\times$) & 0.091\,($2.7\times$) \\
Swin-T & 0.011 & 0.277\,($26\times$) & 0.029\,($2.7\times$) & 0.160\,($15.4\times$) & 0.126\,($11.1\times$) \\
ViT-B/16 & 0.003 & 0.264\,($85\times$) & 0.031\,($8.9\times$) & 0.211\,($64.9\times$) & 0.094\,($23.1\times$) \\
\bottomrule
\end{tabular}
\caption{\textbf{Characteristic-neuron reuse across classes} (mean pairwise top-16 support Jaccard). \emph{Random pair}: two unrelated classes. \emph{Nearest class}: each class and its closest class by feature centroid. \emph{Dogs/Snakes/Birds}: mean within-group Jaccard for the ImageNet dog (151--268), snake (52--68), and bird (7--24) groups; the parenthetical is the ratio to a size-matched random group of classes ($\times$ vs.\ the random pair for the nearest-class column).}
\label{tab:sharing}
\end{table}

\vspace{0.6em}
\noindent
We do not claim that the SAE top-active neurons form a performance-optimal subspace. Rather, weighing this uniform cross-backbone stability and the shared, analyzable structure against a fused near-OOD cost of at most $0.93$ points, we propose them as a deliberate methodological choice for constructing the per-class subspace.